%% file: main.tex
\documentclass[twoside,11pt]{article}
\PassOptionsToPackage{hyperfootnotes=false}{hyperref}
\usepackage[preprint]{jmlr2e}
\usepackage[T1]{fontenc}
\usepackage{lmodern}
\usepackage{lastpage}
\usepackage{amsmath}
\usepackage{array}
\usepackage{booktabs}
\usepackage{xcolor}

\usepackage{framed}
\hypersetup{hidelinks}

\providecommand{\Description}[1]{}
\providecommand{\captionsetup}[1]{}

\definecolor{PromptText}{HTML}{263446}
\definecolor{PromptMuted}{HTML}{5C6B7A}
\definecolor{UserPrompt}{HTML}{175A91}
\definecolor{UserPromptBG}{HTML}{EFF6FC}
\definecolor{PlatformPrompt}{HTML}{BF5414}
\definecolor{PlatformPromptBG}{HTML}{FFF3E9}
\definecolor{UserAgentPrompt}{HTML}{2E723D}
\definecolor{UserAgentPromptBG}{HTML}{EFF8F0}
\definecolor{FeedbackPrompt}{HTML}{684399}
\definecolor{FeedbackPromptBG}{HTML}{F6F0FA}

\newenvironment{promptbox}[3]{%
  \def\PromptAccent{#1}%
  \def\PromptBackground{#2}%
  \def\FrameCommand##1{%
    {\fboxrule=0.5pt\fboxsep=4pt%
    \fcolorbox{\PromptAccent}{\PromptBackground}{##1}}}%
  \MakeFramed{\advance\hsize-\width\FrameRestore}%
  \noindent\colorbox{\PromptAccent}{%
    \strut\hspace{2pt}\textcolor{white}{\sffamily\bfseries\scriptsize #3}\hspace{2pt}}%
  \par\nobreak\smallskip
  \normalcolor\footnotesize\raggedright
  \setlength{\parindent}{0pt}\setlength{\parskip}{1.5pt}%
}{%
  \par\endMakeFramed\smallskip\aftergroup\normalcolor
}

\newcommand{\PromptField}[2]{%
  \par\noindent{\color{PromptMuted}\sffamily\bfseries\scriptsize #1}\enspace #2\par}
\newcommand{\PromptRule}[1]{%
  \par\noindent\hangindent=1em\hangafter=1%
  \textcolor{\PromptAccent}{\textbullet}\hspace{0.45em}#1\par}
\newcommand{\PromptVariant}[2]{%
  \par\noindent{\color{\PromptAccent}\sffamily\bfseries\scriptsize #1}\enspace #2\par}
\newcommand{\PromptSchema}[1]{%
  \par\smallskip\noindent
  \begingroup\setlength{\fboxsep}{3pt}%
  \colorbox{white}{\parbox{\dimexpr\linewidth-2\fboxsep\relax}{%
    \color{PromptText}\ttfamily\scriptsize #1}}%
  \endgroup\par}

\jmlrheading{--}{2026}{1--\pageref{LastPage}}{}{}{}{Deyao Hong, Kehan Zheng,
  Qian Li, Jun Zhang, Jie Jiang, and Hongning Wang}
\ShortHeadings{The User Asks, Platforms Compete}{Hong et al.}
\firstpageno{1}

\title{The User Asks, Platforms Compete:
  How Agentic Recommendation Markets Take Shape}

\author{
\name {\small Deyao Hong$^{1}$\thanks{Equal contribution.},
  Kehan Zheng$^{1}$\footnotemark[1], Qian Li$^{2}$, Jun Zhang$^{2}$,
  Jie Jiang$^{2}$, Hongning Wang$^{1}$\thanks{Corresponding author.}}\\
\addr \makebox[\textwidth][c]{$^{1}$Tsinghua University \qquad $^{2}$Tencent Inc.}\\
{\small\itshape
  \makebox[\textwidth][c]{hongdy22@mails.tsinghua.edu.cn,\quad
    zhengkh24@mails.tsinghua.edu.cn,\quad kathieqli@tencent.com}\\
  \makebox[\textwidth][c]{neoxzhang@tencent.com,\quad zeus@tencent.com,\quad
    hw-ai@tsinghua.edu.cn}}
}

\begin{document}

\maketitle

\begin{abstract}
Online recommendation has traditionally taken place after a user enters a
platform, which determines the candidate pool and the ranking shown to the
user. LLM-based user agents enable a different recommendation process: a user
specifies a need before choosing a platform, leaving platforms to compete for
the user's attention, which we refer to as an \emph{agentic recommendation market}.
In our controlled LLM-based experiments across three product domains, we find
this new setting of recommendation creates a tension between access and attention. Compared
with traditional platform-centric recommendation, user-centric recommendation
greatly expands the opportunity for relevant items to enter comparison; yet
broader participation does not translate directly into effective exposure.
Competition directly triggers platforms' strategic play:
selectively
positive explanations occupy $73$--$78\%$ first-ranked positions. When the
user agent relates platforms' actions to subsequent user feedback, this
share falls to $36$--$41\%$, while the chance of user purchasing the relevant
item increases. A user agent is therefore more than a ranker over a larger
pool of candidates: its querying, ranking, and feedback mechanism governing who can compete, how
scarce attention is allocated, and how earlier outcomes shape the evaluation of
platforms directly affect user utility. Designing agentic recommendation therefore requires treating
access, attention, and accountability as a joint mechanism design problem.
\end{abstract}


\input{sections/introduction}
\input{sections/related-work}
\input{sections/method}
\input{sections/experiments}
\input{sections/discussion}
\input{sections/conclusion}

\appendix
\input{sections/supplementary}
\input{sections/appendix}

\bibliography{references}

\end{document}

%% file: sections/introduction.tex
\section{Introduction}
\label{sec:introduction}

Online recommendation has traditionally taken place after a user enters a platform,
such as YouTube or TikTok. The platform then determines the candidate pool and
the ranking shown to the user
\cite{adomavicius2005toward}. This process is so common that it is often
 ignored when the recommendation problem is formulated. Now, agents from large language models
(LLM) change the story: on the user's behalf, they interpret open
needs, maintain the user context, and coordinate recommendations across different platforms
\cite{huang2025recommender,peng2025survey}. Recent work has explored
user-oriented recommendation agents \cite{xu2025iagent}, user-centric and
governable personalization \cite{zhang2026usercentric,liu2026governable},
cross-platform use of personal data \cite{lin2026usergoverned}, and autonomous
buyers in programmable marketplaces \cite{allouah2025agentbuying}. Together,
these developments present a different recommendation paradigm:
a user specifies a need before choosing a platform, leaving platforms to compete
for the user's attention.

\begin{figure}[t]
  \centering
  \par\kern-3pt\noindent
  \includegraphics[width=\columnwidth]{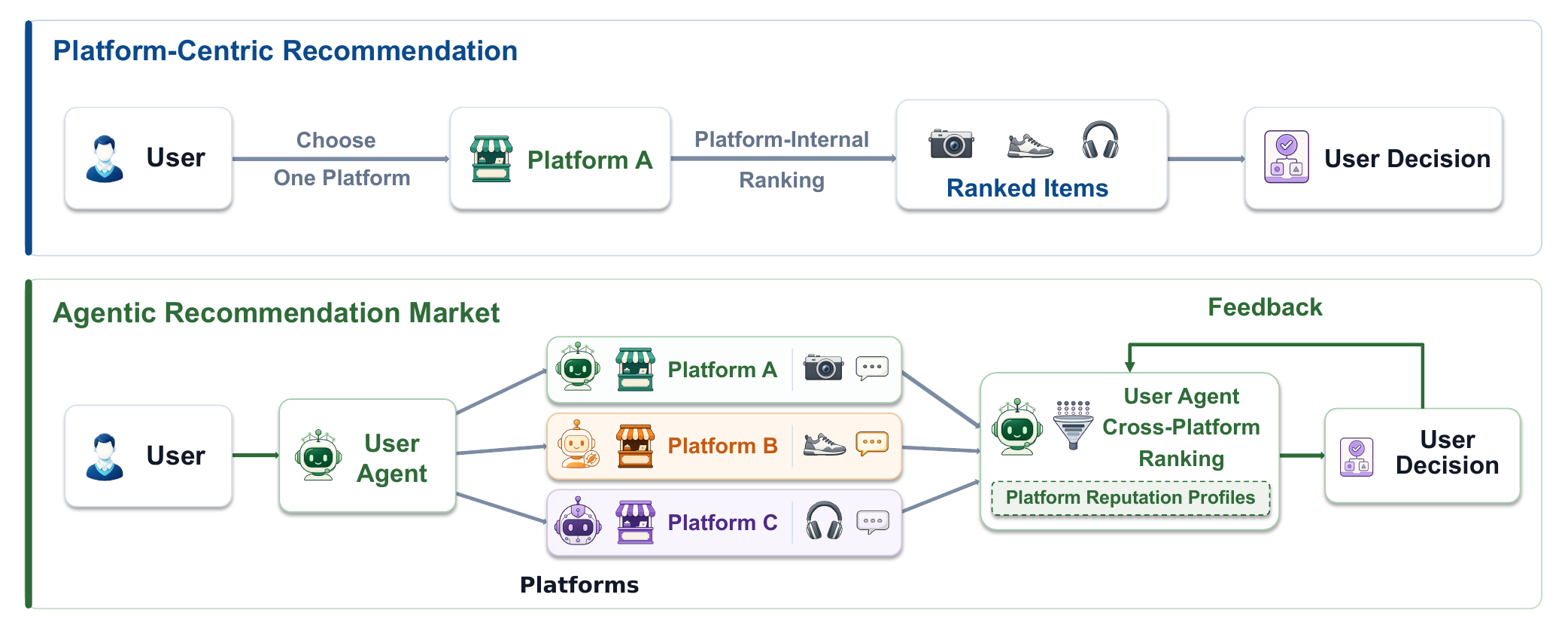}
  \Description{The top panel shows platform-centric recommendation: a user
  chooses one platform, which ranks items before the user decides. The bottom
  panel shows an agentic recommendation market: a user agent queries multiple
  platforms, aggregates their items and recommendation explanations, optionally uses platform
  reputation for cross-platform ranking, and updates reputation from user
  feedback.}
  \captionsetup{aboveskip=-1pt,belowskip=-2pt}
  \caption{From platform-centric recommendation to cross-platform user agent
  mediation.}
  \label{fig:intro-framework}
\end{figure}

We call this new paradigm an \emph{agentic recommendation market}.
Figure~\ref{fig:intro-framework} contrasts it with \emph{platform-centric}
recommendation, in which the user chooses one platform before item-level
comparison. We use the term \emph{user agent} to denote a software intermediary
that interprets a user's request, queries participating platforms, and ranks
recommendation candidates on the user's behalf. Upon the user agent's request, each platform returns a set of
candidate items with corresponding explanations.
The user agent then consolidates the results to construct a
cross-platform ranked list for the user's final decision, i.e., forming user utility. Afterwards,
the user's post-purchase feedback informs user agent's future actions. 

Moving the control of exposure, therefore influence on user's attention, from individual platforms to a user agent creates competition, which in turn brings new challenges in recommendation. 
In this new setting, as the user agent only knows the target user's assessment on the past recommendation candidates, collaborative filtering type features are missing; as a result, the agent can only count on its world knowledge (e.g., in the pretrained models) and platforms' provided explanations to filter and rank the candidates for the user. This creates room for platforms to play \emph{strategically}: they might manipulate the returned set of items and their explanations to drive the user agent's decision in their favor. 
We attribute this new challenge to three inter-dependent aspects: \emph{access}, \emph{attention}, and \emph{accountability}. \emph{Access} asks who can participate: cross-platform querying can discover available item  beyond a precommitted catalog, but availability alone does not guarantee exposure \cite{curmei2021quantifying}. 
\emph{Attention} asks who gets users' attention in the final recommendation: as more platforms participate, the candidate pool can easily grow faster than the positions a user can examine, making the agent the direct influencer of user's scarce attention \cite{singh2018fairness,abdollahpouri2020multistakeholder}. 
\emph{Accountability} asks how platforms can be held accountable regarding their provided content: as platforms can influence how the user agent presents recommendation candidates, their current proposals should be interpreted against user feedback on earlier outcomes \cite{qin2024beyond,dellarocas2003digitization}. 
Together, these aspects connect the life cycle of agentic recommendation market---entering the competition, receiving attention, and carrying feedback into later rounds---rather than three independent objectives.

These new challenges motivate three research questions: 
\begin{itemize}
    \item \textbf{RQ1.} \emph{How does cross-platform solicitation change user utility relative to platform-centric recommendation setting?}
    \item \textbf{RQ2.} \emph{As market participation and user agent controlled exposure vary, how are such resources converted into attention and purchase---and where are they lost?}
    \item \textbf{RQ3.} \emph{How does platforms' strategic explanation framing affect user attention and purchase, and how can evidence from earlier outcomes reshape that competition?}
\end{itemize}

We investigate these questions in a controlled LLM testbed, drawing on recent
uses of generative agents to simulate recommendation interactions
\cite{zhang2024generative,jin2025recinter}. The testbed is grounded in the scenario of item recommendation constructed out of three categories, i.e., musical
instruments, video games, and sports \& outdoors, from Amazon reviews
\cite{hou2026bridging}. Each session reconstructs a shopping request
from a user's interaction history and uses the final highly rated item as a
held-out positive reference (the target). We vary market participation, shortlist capacity,
platform presentation policy, and the availability of platform-specific outcome
histories, tracing the target through candidate-pool presence, shortlist
inclusion, Top-1 attention, and purchase.

\textbf{Study objective and scope.}
Our focus is the market mechanism rather than user agent optimization or
ranking-algorithm comparison. Specifically, we ask who can participate, how 
capacity-constrained exposure allocates attention, and how outcome histories
affect how current platform claims should be treated. 
The testbed operationalizes
the structural contrast between platform precommitment and cross-platform
competition along three dimensions: participation, attention allocation, and
the use of outcome history. Because within-platform logs cannot reveal this
counterfactual cross-platform process, controlled experiments let us trace how
these structural differences propagate from proposal through purchase. We use
the resulting patterns to characterize how the shift from a single-platform
pipeline to a cross-platform market changes access, attention allocation, and
accountability.

Our experiments yield three findings. First, cross-platform
querying changes the opportunity set. Relative to platform precommitment,
the target reaches the candidate pool in roughly one fifth of episodes under
platform-centric recommendation and nearly nine tenths under cross-platform
querying. The stage-wise trace shows that this separation arises before
ranking, because an item outside the precommitted catalog cannot enter
competition. This isolates the upstream access difference from downstream
attention allocation.

Second, broader access creates a downstream attention bottleneck. In the
market-expansion conditions we test, adding platforms makes the target more
likely to enter the candidate pool, yet less likely to enter the shortlist or be purchased.
Increasing shortlist capacity substantially raises shortlist inclusion, but
mainly through lower-ranked positions, with little change in top-ranked
attention or purchase. Opening the market therefore does not remove scarcity;
it moves scarcity from catalog access to the user agent's allocation of attention.

Third, platform presentation becomes part of the competition. We define
\emph{exaggeration} as a presentation policy that uses confident, selectively
positive language and may downplay limitations without inventing objective product
attributes. To test whether outcome feedback can calibrate the influence of
such framing, we compare two otherwise identical conditions. In \emph{NoRep},
the user agent ranks current proposals without platform-specific outcome
histories. In \emph{LocalRep}, it additionally retains and consults
user-specific records of whether each platform's earlier claims were borne out
by post-purchase feedback. With equal supply from three policy families,
exaggeration captures $73$--$78\%$ of the user agent's top-ranked positions
under \emph{NoRep}. Under \emph{LocalRep}, this share falls to $36$--$41\%$,
and target purchase increases in every tested mixed-market composition.
Relating current claims to outcome-grounded histories can therefore change the
basis on which scarce attention is allocated.

Taken together, this work makes three contributions:
\begin{itemize}
  \item \textbf{Market formulation.} We formalize cross-platform user agent
  mediation as a recommendation market and organize its design around Access,
  Attention, and Accountability: who can compete, which recommendation proposals receive attention, and
  how prior outcomes bear on current claims.

  \item \textbf{Controlled empirical testbed.} We build an LLM-based
  market environment grounded in real interaction histories, vary the
  market rules systematically, and measure their stage-wise effects with
  explicit information boundaries.

  \item \textbf{Empirical evidence on market structure.} Across three product domains, we
  document a structural access gain from cross-platform querying, an
  attention bottleneck as supply expands, a gap between display capacity and
  top-ranked attention, and evidence that linking current claims to recorded
  outcomes can direct scarce attention toward platforms with better-supported
  claim histories.
\end{itemize}

%% file: sections/related-work.tex
\section{Related Work}
\label{sec:related-work}

\subsection{Agentic Recommendation Paradigms}

LLM-powered agents are extending recommendation beyond one-shot scoring toward
systems that retain memory, reason over user intent, invoke tools, and interact
over time. Surveys organize this landscape by agent roles and degrees of
autonomy \cite{peng2025survey,lin2026roadmap}. Within a platform, InteRecAgent
orchestrates information querying, retrieval, and ranking, while AgenticRS
recasts pipeline modules as independently evaluable and evolvable agents
\cite{huang2025recommender,hu2026agenticrs}. Simulation-oriented work uses
generative user agents for longitudinal evaluation and merchant agents to create
evolving recommendation environments, while AgentRecBench provides an
interactive textual environment and modular evaluation framework
\cite{zhang2024generative,jin2025recinter,shang2025agentrecbench}. These
directions establish orchestration and simulation capabilities, but remain
centered on operating or evaluating a single recommender system.

Agency is also shifting toward users and providers. iAgent places a
user-oriented intermediary between a user and a recommender, and related work
explores user-centric agents, governable representations, and cross-platform
personalization
\cite{xu2025iagent,zhang2026usercentric,liu2026governable,lin2026usergoverned}.
On the provider side, AgentCF models user and item agents; Rec4Agentverse envisions
Agent Items interacting with an Agent Recommender; and TriRec lets item agents
self-promote before centralized reranking
\cite{zhang2024agentcf,zhang2024rec4agentverse,gong2026trirec}. Together, these
systems instantiate pieces of an agent-mediated ecosystem, but focus on
orchestration, personalization, simulation fidelity, or stakeholder utility
within a recommender. We instead study competing platforms with separate
catalogs: they submit candidate items and recommendation explanations to a user
agent, compete for positions in a capacity-constrained cross-platform shortlist,
and have subsequent claims
evaluated against outcome-grounded histories. Our unit of analysis is the market
protocol through which access, attention, and accountability are distributed.

\subsection{Competition and Accountability in Recommendation Markets}

Recommendation is increasingly understood as a multistakeholder allocation
problem: a ranked list serves users while distributing scarce exposure among
providers \cite{abdollahpouri2020multistakeholder,singh2018fairness}. Work on
reachability further distinguishes content availability from a user's
opportunity to discover it \cite{curmei2021quantifying}. A
complementary game-theoretic literature studies mediators selecting among
strategic providers \cite{benporat2018game}, creators adapting their content
under top-$K$ competition \cite{yao2023howbad}, and supply-side equilibria
induced by personalized recommendation \cite{jagadeesan2023supply}. Related
work on platform competition instead places the user's choice among
recommenders before item-level recommendation and shows that competition need
not align market outcomes with user utility \cite{jagadeesan2023competition}.
More broadly, research on agentic markets studies electronic markets in which AI
agents search, evaluate, negotiate, or transact on behalf of buyers or sellers
\cite{bichler2026agenticmarkets}. Our setting brings these questions together:
within each request, a user agent queries multiple platforms and allocates a
capacity-constrained cross-platform shortlist. Market competition therefore
governs both item availability and downstream attention.

Natural-language recommendation explanations add a signaling channel beyond item
characteristics.
Recommender explanations serve potentially conflicting goals, including
transparency, trust, decision effectiveness, and persuasion
\cite{tintarev2012evaluating}. With LLMs, persuasive explanations may contain
unsupported information \cite{qin2024beyond}, and controlled user studies
show that persuasive wording can increase choices of lower-utility options
\cite{rahman2026persuasive}. Most closely related, ACES shows that a seller can
gain market share by modifying product descriptions for AI shoppers
\cite{allouah2025agentbuying}. In our setting, multiple platforms generate
query-specific recommendation explanations before cross-platform ranking, and we
separate user agent selection from the user's final purchase. Repeated interaction
also creates a role for accountability. Research on online markets shows that
reputation systems summarize transaction histories to reduce uncertainty and
inform future decisions
\cite{dellarocas2003digitization,resnick2006value}. In our setting, we ask
whether simulated post-purchase evaluations can be used to construct
platform-specific reliability records that help the user agent calibrate confidence in
a platform's later explanations.

%% file: sections/method.tex
\section{The Agentic Recommendation Market}
\label{sec:market}

The market comprises a set of users, a \emph{user agent} that queries and compares
recommendations on each user's behalf, and $N$ competing platforms, each with a catalog
$\mathcal{I}_p$. Rather than choosing a
platform first, the user presents the need through the user agent to the platforms
$\mathcal{P}=\{p_1,\ldots,p_N\}$. Each platform returns a set of candidate items and corresponding
recommendation explanations; the user agent
returns a ranked shortlist of capacity $1\le K\le N$, and the user makes the
purchase decision. The protocol thereby separates the opportunity to compete
from the scarce exposure allocated through the shortlist.

\begin{figure}[t]
  \centering
  \includegraphics[width=.96\textwidth]{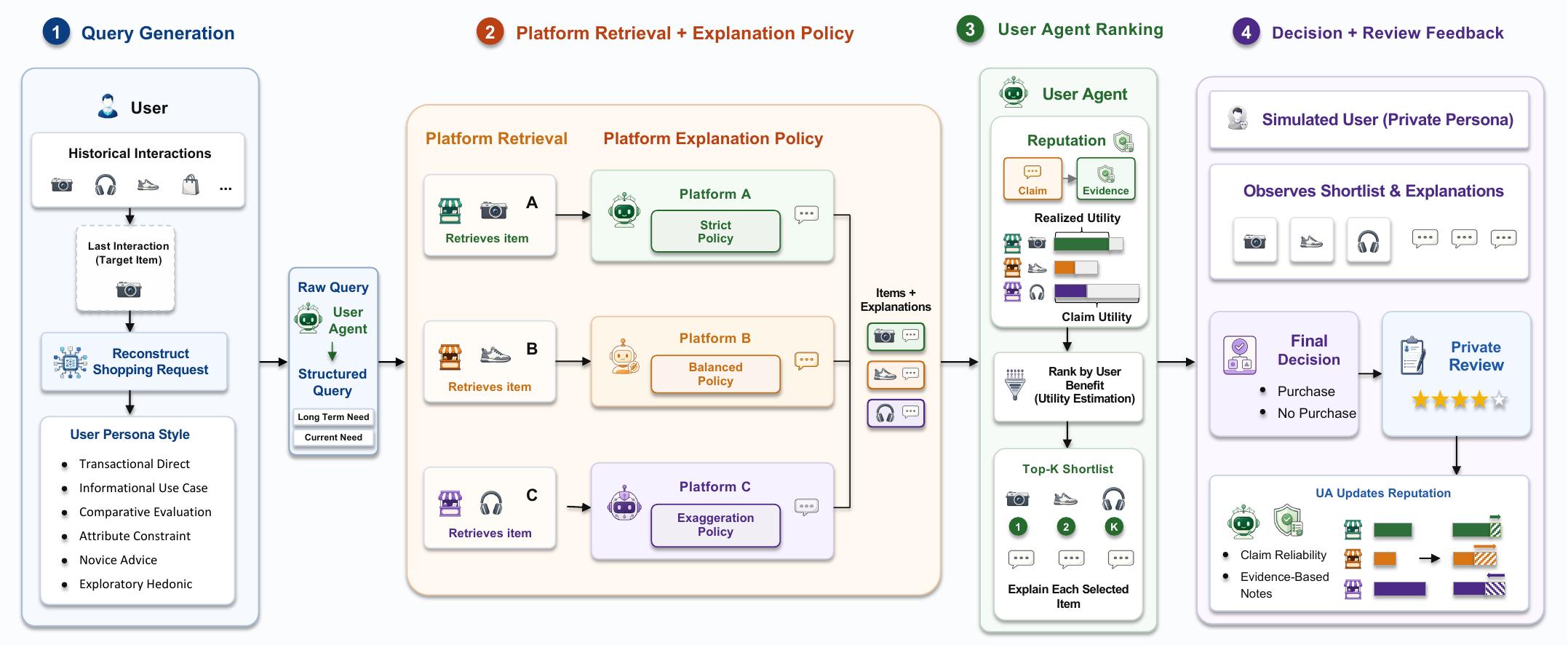}
  \Description{The market interaction has four stages. A simulated user query
  is generated from prior interactions and a held-out target, then structured
  by a user agent. Each platform retrieves one item and writes a
  recommendation explanation. The user agent ranks the proposals into a
  top-$K$ cross-platform shortlist, optionally using platform reputation. The
  user then purchases an item or declines, and post-purchase feedback can update
  the selected platform's reputation.}
  \caption{A round in an agentic recommendation market. The user agent queries
  candidate items and explanations from multiple platforms, constructs a capacity-constrained
  cross-platform shortlist, and can use post-purchase feedback to update
  user-scoped platform reputation.}
  \label{fig:market-framework}
\end{figure}

\subsection{Market Roles and Information Boundaries}

The user expresses demand and makes the final choice; the user agent structures that
demand and allocates shortlist attention; and platforms retrieve and present
candidate items.

\paragraph{User.}
The user has an interaction history, an immediate demand, and a private decision
style hidden from other participants. The same private style shapes request
generation, final
choice, and (when applicable) post-purchase evaluation; these are facets of one
user role, not separate market actors. The user expresses a need, then observes
the shortlist, platform recommendation explanations, and user agent rationales before
purchasing or declining.

\paragraph{User Agent.}
The user agent converts the request and observable history into a structured need,
sends it to all platforms, and ranks their proposals, optionally consulting
local platform histories. It owns no catalog, writes no platform explanation, and does not
make the purchase decision. It receives neither the user's private style nor
the held-out target identity.

\paragraph{Platforms.}
Each platform uses only its catalog and visible item evidence to select one
candidate and generate a recommendation explanation under a preassigned
presentation policy. It cannot
observe other proposals, the private user style, or the target label.
Presentation-policy and user-style
controls appear in Appendix~\ref{app:style-design}; core prompts appear in
Appendix~\ref{app:prompts}.

\paragraph{Constructing a recommendation episode.}
Offline logs record what a user selected and reviewed, but not the preceding
request or a counterfactual evaluation of another purchase. We hold out the
user's final interaction with a rating of at least four and use the preceding history
to reconstruct a consistent episode: history supplies longer-term context,
while the held-out
rating and review describe one favorably received outcome. The same private
evidence is used to construct a plausible request for which the target is a
strong but non-unique match and, after a purchase, to generate feedback on the
selected item and recommendation explanation. This information is withheld from the platforms, user agent
ranker, and final decision model; feedback occurs only after that choice.

\subsection{Market Interaction Protocol}

Figure~\ref{fig:market-framework} summarizes one round of interaction in four stages.

\paragraph{Stage 1: Query generation and structuring.}
Let $i_u^*$ denote the held-out positive reference, which we call the target,
and let $H_u$ denote the preceding history. The user's request-generation
component produces the reconstructed request $q_u$ described above. The user agent then
forms a structured representation
\begin{equation}
  z_u=g_{\mathrm{user\,agent}}(H_u,q_u)
      =\bigl(z_u^{\mathrm{long}},z_u^{\mathrm{current}}\bigr),
\end{equation}
where $z_u^{\mathrm{long}}$ captures longer-term preferences inferred from
history and $z_u^{\mathrm{current}}$ captures the immediate need. This same
representation is sent to every platform. Neither the user agent, the platforms, nor
the final decision model is told which item is the target; if retrieved,
$i_u^*$ appears with the same visible fields as any other candidate.

\paragraph{Stage 2: Platform retrieval and explanation generation.}
Each platform retrieves one item from its own inventory,
\begin{equation}
  i_{u,p}=\mathcal{R}_p(z_u,\mathcal{I}_p),
\end{equation}
and generates a query-specific recommendation explanation under its fixed
presentation policy $\sigma_p$,
\begin{equation}
  m_{u,p}=\mathcal{G}_p(z_u,i_{u,p};\sigma_p).
\end{equation}
The platform proposal is $c_{u,p}=(p,i_{u,p},m_{u,p})$, and all proposals
form the cross-platform candidate pool
$\mathcal{C}_u=\{c_{u,p}:p\in\mathcal{P}\}$. Each platform receives exactly one
proposal opportunity. If an item is carried by multiple platforms, the
proposals remain distinct because their explanations and platform histories may
differ.

\paragraph{Stage 3: cross-platform ranking by the user agent.}
The user agent jointly considers the explicit request, its structured understanding,
visible item evidence, and the current recommendation explanations. When enabled, it also observes
the platform-specific histories maintained for the current user agent--user
relationship, denoted $\mathcal{H}_{s(u),p}$ and defined in
Section~\ref{sec:localrep}. It ranks candidates by apparent fit to the user's
request and produces
\begin{equation}
  L_u=\pi_{\mathrm{user\,agent}}
  \bigl(q_u,z_u,\mathcal{C}_u,
  \{\mathcal{H}_{s(u),p}\}_{p\in\mathcal{P}}\bigr),
  \qquad |L_u|=K.
\end{equation}
For every selected item, the user agent provides a user-visible rationale for its
inclusion. The user agent does not decide whether a transaction occurs; its output is
only the ranked shortlist. Candidate-pool inclusion therefore represents
market access, whereas shortlist inclusion represents effective attention.

\paragraph{Stage 4: User decision and feedback.}
The user observes the shortlist, platform recommendation explanations, and user agent rationales, and
chooses an action
\begin{equation}
  a_u\in L_u\cup\{\varnothing\},
\end{equation}
where $a_u=\varnothing$ denotes no purchase. When feedback is enabled and a
purchase occurs, the post-purchase component generates a simulated review
$v_u$ of the selected item. The reputation updater uses this review to assess
whether the selected platform's pre-purchase claims were borne out by the
simulated experience and stores that assessment in the user-scoped platform
history. In later rounds, the user agent can use this history to calibrate its confidence
in the platform's current claims when evaluating new platform recommendations.

\subsection{Outcome-Grounded Local Reputation}
\label{sec:localrep}

In the market architecture we study, every user is assisted by their own user agent.
Knowledge acquired by one user agent is scoped to its continuing relationship with that
user and is not automatically shared with other users' user agents. We
refer to this user-scoped platform history as \emph{LocalRep}. It gives the user agent
a simple way to hold platforms accountable across interactions: claims made
before one purchase can be checked against the user's subsequent experience
and can then inform how the same user agent interprets future platform recommendations. Let $s(u)$ denote
the continuing user agent--user relationship associated with episode $u$. For every
platform $p$, that user agent maintains a bounded history $\mathcal{H}_{s(u),p}^{(t)}$ of recent
claim-reliability records.

When the user purchases from platform $p^*$, the reputation updater compares
the selected platform's pre-purchase recommendation explanation $m_{u,p^*}$ with the simulated
post-purchase review $v_u$, using the observable purchase context $x_u$, and
produces a claim-reliability entry $e_u^{(t)}$. The selected platform's user-scoped
history is then updated as
\begin{equation}
  \begin{aligned}
    e_u^{(t)}
      &=F_{\mathrm{rep}}\!\left(m_{u,p^*},v_u;x_u\right),\\
    \mathcal{H}_{s(u),p^*}^{(t+1)}
      &=\operatorname{Tail}_{W}\!\left(
        \mathcal{H}_{s(u),p^*}^{(t)}\oplus e_u^{(t)}\right),
  \end{aligned}
\end{equation}
where $\operatorname{Tail}_{W}$ retains the $W$ most recent entries and
$\oplus$ denotes sequence concatenation. Each entry contains a reliability
judgment and a short evidence note. Other platform histories remain unchanged, and a round ending
without a purchase produces no update. Complete updater inputs and prompts
appear in Appendix~\ref{app:prompt-accountability}.

The updater asks whether the experience supports the expectations created by
the recommendation explanation; it does not equate product satisfaction with
claim reliability. A
candidly disclosed limitation need not yield a lower-reliability record when
the review confirms that trade-off. An omitted or understated core limitation, by
contrast, can produce a lower-reliability entry when it creates an expectation
gap.

In later rounds, the user agent uses these records to calibrate how it interprets each
platform's current claims. LocalRep thereby links current claims to demonstrated
reliability, allowing shortlist attention to reflect the platform's historical
claim calibration. The NoRep condition retains the same
market flow but omits both the reading and updating of these histories.

%% file: sections/experiments.tex
\section{Experiments}
\label{sec:experiments}

We evaluate the market along the three functions introduced in
Section~\ref{sec:introduction}: expanding access to cross-platform candidates,
allocating limited user attention, and linking platform claims to simulated
post-purchase outcomes. We first compare \emph{Platform-Centric} with user agent-mediated
recommendation, then examine how market size, shortlist capacity, platform-side
explanation policies, and outcome-grounded reputation affect recommendation
outcomes.

\subsection{Experimental Setup}
\label{sec:exp-setup}

\paragraph{Datasets and episode construction.}
We use Musical Instruments, Video Games, and Sports \& Outdoors from Amazon
Reviews'23 \cite{hou2026bridging}. Within each domain, interactions are ordered
chronologically. We retain users with at least five interactions, keep at most
the 20 most recent, and hold out the final interaction with a rating of at least
four
as the positive reference $i_u^*$; the remaining interactions form the
observable history. We generate 1,200 requests per domain, balanced across the
six profiles in Appendix~\ref{app:user-profiles}. Only the query generator has
access to the held-out interaction before purchase, allowing it to reconstruct a
plausible need without copying or uniquely identifying the item; platforms, the
user agent, and the final decision model never receive its target status. We trace
$i_u^*$ through submission, shortlisting, Top-1, and purchase. Target Purchase
records whether this held-out positive reference is ultimately selected; an
off-target purchase is a purchase of another shortlisted item, not an evaluation
error. Post-purchase use of the reference is documented in
Section~\ref{sec:market}.

\paragraph{Market conditions.}
Each episode contains $N$ platform catalogs; each platform returns one candidate
item and recommendation explanation, and the user agent returns a shortlist of capacity
$K$. Each item is assigned to a primary platform and copied independently to every
other platform with
probability $\rho$. Catalogs contain at most $C$ regular items while retaining
their assigned targets. The default heterogeneous market assigns equal platform
shares to Strict, Balanced, and Exaggeration. These presentation policies emphasize
evidence and limitations, balance supported benefits with trade-offs, or use
confident, selectively positive framing without inventing objective attributes,
respectively. Appendix~\ref{app:platform-policies} gives full definitions.

\emph{Platform-Centric} operationalizes platform precommitment: the user enters
one uniformly sampled platform and receives $J$ items from its catalog. Uniform
entry is appropriate because the constructed platforms are ex ante
exchangeable, with no brand, specialization, or observed entry preference. By
contrast, user agent-mediated conditions query all participating catalogs before
forming a capacity-$K$ shortlist. All conditions use the same requests, catalogs,
retrieval, item evidence, and user profiles, with $J=K$; only the timing and
scope of platform choice differ. \emph{user agent--NoRep} omits histories, whereas
\emph{user agent--LocalRep} retains up to $W$ records per platform and updates only
the platform from which the purchase was made. Section~\ref{sec:localrep} and
Appendices~\ref{app:style-controls} and~\ref{app:prompts} give the update rule,
prompts, and information boundaries.

\paragraph{Controlled target opportunity.}
To isolate downstream competition from retrieval, we control the target's
opportunity to enter the first-stage candidate pool. For each platform whose
catalog contains $i_u^*$, the simulation runner includes it among that
platform's candidate items
with probability $p_{\mathrm{hit}}$; otherwise, candidates come from its top-$L$
embedding-retrieval results and may still include it. This catalog-conditional
rule applies to the entered platform under Platform-Centric and independently
across queried catalogs under user agent mediation; it never adds the target to a
catalog that does not carry it. Conditional on $m$ of the $N$ catalogs carrying
the target, the forced-inclusion component makes the target candidate
available with probability $(m/N)p_{\mathrm{hit}}$ under Platform-Centric and
$1-(1-p_{\mathrm{hit}})^m$ under user agent mediation. The latter probability is
weakly larger. This inequality captures the structural access difference under
user agent mediation: by carrying the user's request across platform boundaries, the user agent
can surface a relevant item from any participating catalog that carries it,
rather than leaving access contingent on which platform the user entered first.
Natural retrieval may additionally recover the target in either condition; the full
derivation appears in Appendix~\ref{app:controlled-access}. Target Present,
Shortlist, Top-1, and Purchase then show how candidate-pool access is converted
into attention and purchase.

\begin{figure}[!t]
  \centering
  \includegraphics[width=.89\textwidth]{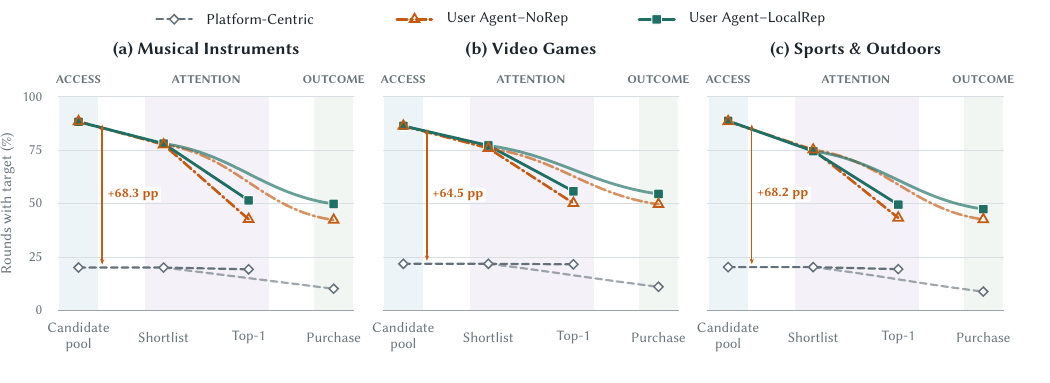}
  \Description{Three domain panels trace the target through candidate-pool
  access, shortlist inclusion, Top-1 attention, and purchase for a
  Platform-Centric reference condition, user agent--NoRep, and user agent--LocalRep. Opening the
  market produces a large increase in candidate-pool access; LocalRep mainly
  changes downstream ranking and purchase.}
  \captionsetup{aboveskip=-8pt,belowskip=0pt}
  \caption{Stage-wise target outcomes under Platform-Centric and user agent-mediated
  recommendation ($J=K$).}
  \label{fig:access-stagewise}
\end{figure}

\begin{figure}[!t]
  \centering
  \includegraphics[width=.89\columnwidth]{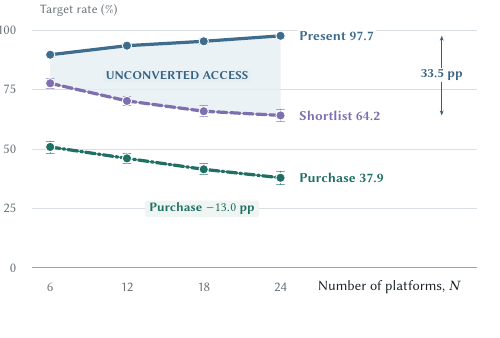}
  \Description{As the number of platforms increases from six to twenty-four,
  mean target presence rises, while mean shortlist inclusion and target
  purchase fall. The shaded gap between presence and shortlist inclusion
  widens.}
  \captionsetup{aboveskip=-11pt,belowskip=0pt}
  \caption{Market expansion increases target availability but intensifies
  attention competition. Lines show domain means; whiskers show 95\%
  confidence intervals.}
  \label{fig:attention-market-size}
\end{figure}

\begin{figure}[!t]
  \centering
  \includegraphics[width=.89\columnwidth]{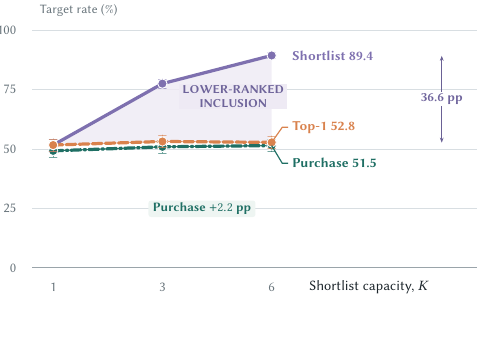}
  \Description{As shortlist capacity increases from one to six, target
  shortlist inclusion rises sharply, but target Top-1 attention and purchase
  remain near one half. The shaded gap indicates added lower-ranked exposure.}
  \captionsetup{aboveskip=-8pt,belowskip=1pt}
  \caption{Shortlist expansion mainly adds lower-ranked exposure. Conventions
  follow Figure~\ref{fig:attention-market-size}.}
  \label{fig:attention-shortlist-capacity}
\end{figure}

\begin{figure}[!t]
  \centering
  \includegraphics[width=.96\textwidth]{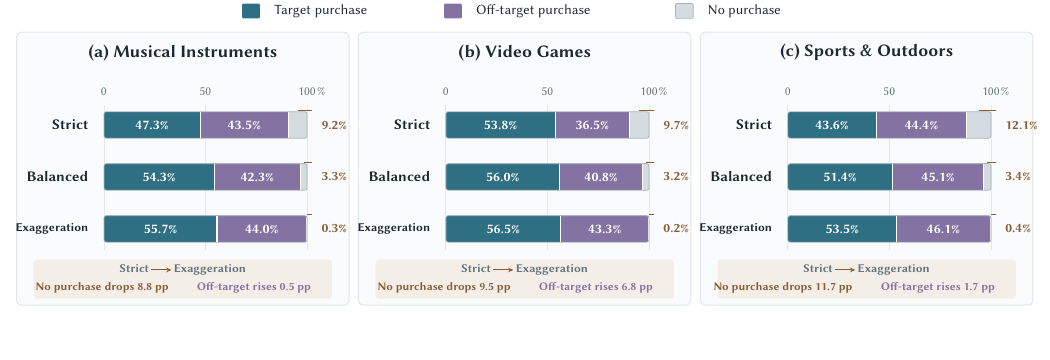}
  \Description{For each domain, three one-hundred-percent stacked bars compare
  target purchase, off-target purchase, and no purchase in homogeneous Strict,
  Balanced, and Exaggeration markets under NoRep. Exaggeration sharply
  reduces no purchase and also raises off-target purchase relative to Strict.}
  \captionsetup{aboveskip=-8pt,belowskip=0pt}
  \caption{Decision composition under homogeneous NoRep explanation policies.
  Callouts compare Strict with Exaggeration. ``Off-target'' means purchase of a
  shortlisted alternative to the held-out positive reference.}
  \label{fig:signaling-outcomes}
\end{figure}

\paragraph{Default configuration and treatments.}
Defaults are $N=6$, $K=3$, $C=30{,}000$, $\rho=0.1$,
$p_{\mathrm{hit}}=0.8$, $L=5$, and $W=10$; Platform-Centric uses $J=3$.
We simulate 12 users---two per profile---with 1,200 episodes per domain for the
main comparison and 600 for each additional treatment.

User agent--LocalRep sweeps $N\in\{6,12,18,24\}$ at fixed $K=3$ and
$K\in\{1,3,6\}$ at fixed $N=6$. Because the source item universe is fixed,
changing $N$ reallocates items across catalogs and changes competing candidate
items,
target multiplicity, and feedback density; we therefore interpret it as a
market-structure treatment rather than an isolated platform-count effect.
Signaling uses homogeneous Strict, Balanced, or Exaggeration markets under
NoRep; accountability uses an equal three-policy market and two evenly split
two-policy markets. Appendix~\ref{app:experiment-matrix} gives the full matrix
and exact policy shares.

We report Target Present, Shortlist, Top-1, and Purchase; purchase may occur at
any shortlist rank.

\paragraph{Statistical analysis.}
For contrasts among stateless NoRep conditions, we use a paired bootstrap over
queries. For contrasts involving stateful LocalRep, we instead resample the 12
simulated users as clusters, preserve each user's complete ordered history, and
use the same resampled user in both treatment arms. Reported 95\% confidence
intervals are based on 50,000 draws stratified by domain. Macro estimates are
unweighted means of the three domain estimates. Primary intervals condition on the fixed model, seed,
catalog assignment, and observed episode order; separate robustness experiments
vary catalog assignment and evaluation model. Full comparison sets appear in
Appendices~\ref{app:statistical-analysis} and~\ref{app:robustness}.

\paragraph{Implementation.}
Main experiments use DeepSeek-V4-Flash for all generative roles, and dense
retrieval uses BAAI BGE-base-en-v1.5. Candidates are sampled from the top-$L$
retrieval set with Gumbel temperature $\tau=0.1$; all experiments use seed 42.
The model-sensitivity analysis holds query generation and platform-side explanation
fixed while varying the evaluation-side model, as detailed in
Appendix~\ref{app:robustness}.
\par\noindent Code and experiment materials will be made available at:\\
\href{https://github.com/hongdy22/Simulated-Agentic-Recommendation-Market}{%
\texttt{hongdy22/Simulated-Agentic-Recommendation-Market}}.

\subsection{Cross-Platform Querying Changes Who Can Compete}
\label{sec:exp-access}

Figure~\ref{fig:access-stagewise} traces how the timing of platform choice
changes the path from proposal to purchase. Both conditions show the user the
same number of items ($J=K$). Platform-Centric fills all displayed positions from the
platform entered before item-level comparison, whereas user agent--NoRep queries all
participating platforms for candidate recommendations before forming a shortlist
of the same size. Accordingly, Target Present rises from roughly one fifth of episodes to
nearly nine tenths, with the separation already established before ranking.

This access advantage carries through to purchase: Target Purchase under
user agent--NoRep is $4.2$--$4.8\times$ the Platform-Centric rate. Yet, among episodes
in which the target is present, the two conditions have much closer conditional
target-purchase rates. The aggregate difference therefore originates at Access:
the user agent carries the same request across platform boundaries, allowing candidate
items excluded by a single precommitted catalog to enter competition.

LocalRep exhibits a different stage-wise pattern. It leaves target presence
nearly unchanged but changes downstream ordering and raises Target Purchase.
Figure~\ref{fig:access-stagewise} thus separates two successive mechanisms:
cross-platform querying changes who can compete, while outcome histories shape
how submitted recommendations receive attention.

\subsection{When Access Becomes an Attention Bottleneck}
\label{sec:exp-attention}

\paragraph{Market expansion.}
Figure~\ref{fig:attention-market-size} reveals a structural reversal: as the
number of participating platforms $N$ grows, the target becomes
more likely to enter the candidate pool, yet less likely to enter the
fixed-capacity shortlist or be purchased. From $N=6$ to $N=24$, Target Present rises by about
8.0 percentage points, while shortlist inclusion and Target Purchase fall by
about 13.6 and 13.0 points, respectively.

This reversal does not reflect reduced target supply. The mean number of target
candidate copies more than doubles, but the full pool of submitted candidates
grows still faster, nearly halving the target's share of the pool. Target Present
requires only that at least one target candidate enter the pool; Attention depends
on that candidate's position relative to all competitors. With $K$ fixed, additional
platforms therefore expand Access while intensifying competition for a limited
number of displayed positions. Market expansion can produce more target copies
and less effective exposure at the same time.

\paragraph{Shortlist capacity.}
Figure~\ref{fig:attention-shortlist-capacity} next tests whether adding display
positions relieves this bottleneck. Increasing $K$ from 1 to 6 raises target
shortlist inclusion by about 37.7 percentage points, while Top-1 attention and
Target Purchase change by only about 1.1 and 2.2 points, respectively, with small or
non-monotonic changes across domains.

The new positions primarily admit items that would otherwise remain outside the
shortlist; they do not create another top-ranked position. An item can therefore
move from invisible to visible while remaining lower in the ranking, and the
additional exposure does not translate into a stable increase in Target
Purchase in this testbed.

Together, the market-size and shortlist-capacity experiments show that the
Attention constraint is not reducible to shortlist capacity.
Market expansion makes more candidate recommendations compete for fixed positions, whereas
shortlist expansion primarily adds exposure below Top-1. Once the user agent crosses
catalog boundaries, scarcity moves inside the cross-platform shortlist and is
concentrated most strongly at its top.

\subsection{From Item Competition to Explanation Competition}
\label{sec:exp-signaling}

The preceding experiments concern which items can enter the market and receive
exposure. Presentation adds a second competitive dimension: once a platform can
explain why a candidate item fits the request, its explanation becomes part of what the
user agent evaluates. We therefore construct homogeneous NoRep markets in which all
platforms use Strict, Balanced, or Exaggeration, while retrieval and the
remaining market rules remain fixed.

Figure~\ref{fig:signaling-outcomes} shows that Exaggeration nearly eliminates
no-purchase outcomes in all three domains. Relative to Strict, it raises the
macro total-purchase rate by about 10.0 percentage points. More assertive,
selectively positive framing changes both which item is purchased and whether a
transaction occurs.

Once platforms can generate recommendation explanations, they compete not only
through their items but also through how those items are presented. Promotional
framing can change both whether a transaction occurs and which recommendation
captures it. A market that evaluates platforms only by immediate transaction volume may therefore
conflate persuasive influence with better need matching. The next question is
whether historical outcomes can calibrate the influence of current claims. As a
boundary analysis, Appendix~\ref{app:overlap-persuasion} examines presentation competition when
high catalog overlap makes different explanations for the same item more common.

\subsection{Outcome-Grounded Accountability Reallocates Attention}
\label{sec:exp-accountability}

\begin{figure}[t]
  \centering
  \includegraphics[width=.96\columnwidth]{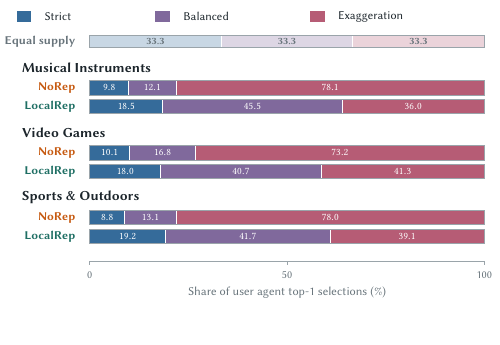}
  \Description{For each domain, paired stacked bars compare the shares of all
  user agent Top-1 selections received by Strict, Balanced, and Exaggeration platforms
  under NoRep and LocalRep. LocalRep sharply reduces the
  Exaggeration share and reallocates attention toward Balanced and Strict.}
  \captionsetup{aboveskip=-8pt,belowskip=0pt}
  \caption{Top-1 attention by presentation-policy family in the equal-supply mixed
  market. Each family controls one third of platforms. Across domains,
  LocalRep reduces Exaggeration's share by 37.62 pp.}
  \label{fig:accountability-attention}
\end{figure}

\paragraph{Reallocating attention.}
The mixed market tests whether outcome history changes the evidence base for user agent
attention allocation when different presentation policies compete directly. With
equal platform supply from all three policy families, Exaggeration captures
73--78\% of Top-1 attention under NoRep, far above its one-third supply share
(Figure~\ref{fig:accountability-attention}). LocalRep lowers this share to
36--41\%, redirecting attention toward Balanced and Strict platforms.

This shift does not require the user agent to identify a platform's policy: neither
the user agent nor the reputation updater receives policy labels. LocalRep records
whether a platform's earlier claims aligned with simulated post-purchase
experience and relates that record to its current claims.

\begin{figure}[t]
  \centering
  \includegraphics[width=.96\columnwidth]{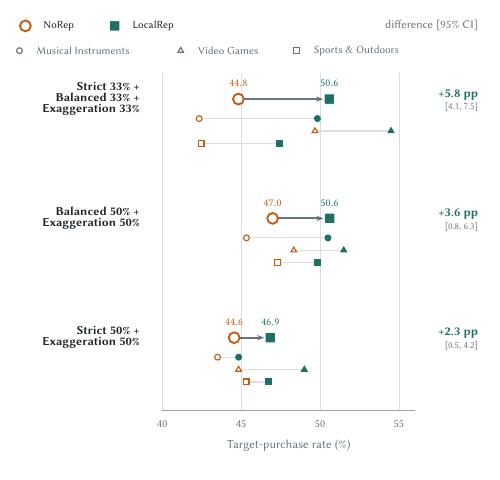}
  \Description{Three dumbbell groups compare target-purchase rates under NoRep
  and LocalRep for an equal three-policy market, a half Balanced and
  half Exaggeration market, and a half Strict and half Exaggeration market. All
  nine domain-level point estimates increase under local reputation.}
  \captionsetup{aboveskip=-8pt,belowskip=0pt}
  \caption{Target-purchase rates under NoRep and LocalRep across mixed-market
  compositions. Large endpoints are unweighted domain means; small markers are
  paired domain estimates. Brackets show 95\% stratified cluster-bootstrap
  confidence intervals for the macro differences.}
  \label{fig:accountability-purchase}
\end{figure}

\paragraph{Downstream purchase.}
The attention shift also reaches the final decision
(Figure~\ref{fig:accountability-purchase}). Across the equal, Balanced--Exaggeration,
and Strict--Exaggeration mixed markets, LocalRep raises macro Target Purchase by
about 5.8, 3.6, and 2.3 percentage points, respectively. All nine paired
domain--composition estimates move in the same direction.

The purchase changes are much smaller than the Top-1 shifts. LocalRep directly
changes the user agent's ordering, whereas the user may still choose from any shortlist
position. Accountability therefore acts most directly by reallocating the
shortlist's single Top-1 position; the change in Target Purchase is a
smaller downstream consequence of that reallocation.

Together with Section~\ref{sec:exp-signaling}, these results show that platform
presentation can dominate attention when current claims are evaluated without
historical evidence. Relating those claims to a platform's demonstrated
reliability can instead direct scarce attention toward platforms with
better-supported claim histories. Cross-platform querying determines which
platforms can participate, the shortlist determines which recommendation
proposals receive attention, and outcome history shapes
how much influence a platform's claims can continue to receive.

\paragraph{Robustness.}
LocalRep's attention reallocation persists under two cyclic inventory
reassignments and four evaluation models, producing a substantial
redistribution of Top-1 attention across policy families in every setting. Target
Purchase also increases in every robustness setting; complete estimates appear in
Appendix~\ref{app:robustness}.

%% file: sections/discussion.tex
\section{Discussion}
\label{sec:discussion}

\subsection{Scope of the Protocol}

An agentic recommendation market is not simply a conventional recommender
applied to a larger catalog. Its querying, shortlisting, and feedback rules move
intermediation from an individual platform's ranking stack to a cross-platform
protocol. These rules should be designed jointly: access determines which
proposals can be compared, whereas exposure and feedback determine which claims
receive attention and later accumulate evidence.

Our Platform-Centric condition isolates precommitment among ex ante symmetric
platforms. The estimates therefore characterize a controlled, short-run protocol
rather than a universal advantage over informed platform choice. In deployment,
user priors, platform participation, and catalog overlap will determine the
access gap. As catalogs converge, competition may shift toward alternative
presentations of the same item (Appendix~\ref{app:overlap-persuasion}).

\subsection{Accountability in a Dynamic Market}

LocalRep is deliberately minimal: it gives the user agent a user-scoped record
of whether earlier recommendation explanations aligned with subsequent
experience. It targets claim calibration rather than product satisfaction alone,
and neither the user agent nor the updater observes presentation-policy labels.

Such evidence is selective, depends on earlier exposure, and may vary across
users, contexts, and time. A deployed mechanism would therefore need to
represent uncertainty, distinguish sparse from unfavorable records, and let new
or improving platforms establish evidence. Platforms, the user agent, and users
may also adapt over time, making both exposure and accountability histories
endogenous. Studying these dynamics is an important next step.

%% file: sections/conclusion.tex
\section{Conclusion}
\label{sec:conclusion}

This paper studies a protocol in which a user agent queries competing platforms for
candidate recommendations. Across three Amazon domains, cross-platform querying
expands target access relative to platform precommitment, but larger markets
intensify shortlist competition, and additional slots mainly add lower-ranked exposure.
Under NoRep, persuasive framing captures disproportionate Top-1 attention;
LocalRep redirects attention toward claims better supported by simulated
outcomes.

A user agent is therefore not merely a larger-catalog ranker. Its querying, display,
and feedback rules jointly govern who can compete, which recommendation
proposals receive scarce attention, and how prior outcomes shape future
influence. Agentic recommendation should treat Access, Attention, and
Accountability as a joint design problem.

%% file: sections/supplementary.tex
\section{Supplementary Experiments and Mechanism Analysis}
\label{app:supplementary}

This appendix consolidates the experimental design and reports analyses of how
reputation scope, accumulated outcome evidence, catalog overlap,
catalog-to-platform assignment, and evaluation-model choice affect the
mechanisms and results studied in the main text. These analyses do not
introduce a new optimization objective: they probe when the reported effects
emerge and what information drives them.

\subsection{Complete Experimental Matrix}
\label{app:experiment-matrix}

Table~\ref{tab:experiment-matrix} records the exact platform composition of every
reported experiment. An equal three-policy market contains two Strict, two
Balanced, and two Exaggeration platforms when $N=6$; because every platform
submits one candidate item, each family represents both 33.3\% of platforms and 33.3\% of
candidate supply. A 50/50 market contains three platforms from each listed
family. Unless varied below, the common settings are $N=6$, $K=3$,
$C=30{,}000$, $\rho=0.1$, $p_{\mathrm{hit}}=0.8$, $L=5$, memory window
$W=10$, and Gumbel temperature $\tau=0.1$.

\begin{table}[t]
  \caption{Complete experimental matrix. Platform composition gives the percentage
  of participating platforms assigned to each presentation policy. ``Local-12''
  denotes two simulated users per operational user profile and is
  the default memory scope used by user agent--LocalRep;
  ``Style-6'' denotes one history per profile; and ``Shared-1'' pools all
  profiles into one reputation history. Rounds are episodes per
  domain and treatment cell.}
  \label{tab:experiment-matrix}
  \small
  \setlength{\tabcolsep}{3.3pt}
  \renewcommand{\arraystretch}{1.12}
  \resizebox{\textwidth}{!}{%
  \begin{tabular}{@{}p{0.125\textwidth}p{0.235\textwidth}
      rrr p{0.125\textwidth}p{0.13\textwidth}r@{}}
    \toprule
    & & \multicolumn{3}{c}{\textbf{Platform composition}} & & & \\
    \cmidrule(lr){3-5}
    \textbf{Experiment} & \textbf{Treatment cells} &
    \textbf{Strict} & \textbf{Balanced} & \textbf{Exaggeration} &
    \textbf{Reputation} & \textbf{Fixed market} & \textbf{Rounds} \\
    \midrule
    Market opening & Platform-Centric; user agent--NoRep; user agent--LocalRep &
    33.3\% & 33.3\% & 33.3\% & None; Local-12 &
    $N=6,K=3,\rho=.1$ & 1,200 \\

    Market expansion & $N\in\{6,12,18,24\}$ &
    33.3\% & 33.3\% & 33.3\% & Local-12 &
    $K=3,\rho=.1$ & 600 \\

    Shortlist capacity & $K\in\{1,3,6\}$ &
    33.3\% & 33.3\% & 33.3\% & Local-12 &
    $N=6,\rho=.1$ & 600 \\

    Homogeneous signaling & All Strict &
    100\% & 0\% & 0\% & None & $N=6,K=3,\rho=.1$ & 600 \\
    Homogeneous signaling & All Balanced &
    0\% & 100\% & 0\% & None & $N=6,K=3,\rho=.1$ & 600 \\
    Homogeneous signaling & All Exaggeration &
    0\% & 0\% & 100\% & None & $N=6,K=3,\rho=.1$ & 600 \\

    Accountability & Equal market: two platforms per policy family &
    33.3\% & 33.3\% & 33.3\% & None; Local-12 &
    $N=6,K=3,\rho=.1$ & 1,200 \\
    Accountability & Balanced--Exaggeration: three platforms per family &
    0\% & 50\% & 50\% & None; Local-12 &
    $N=6,K=3,\rho=.1$ & 600 \\
    Accountability & Strict--Exaggeration: three platforms per family &
    50\% & 0\% & 50\% & None; Local-12 &
    $N=6,K=3,\rho=.1$ & 600 \\

    Catalog assignment & One- and two-step cyclic inventory rotations &
    33.3\% & 33.3\% & 33.3\% & None; Local-12 &
    $N=6,K=3,\rho=.1$ & 600 \\

    Model sensitivity & Four evaluation models on Sports \& Outdoors &
    33.3\% & 33.3\% & 33.3\% & None; Local-12 &
    $N=6,K=3,\rho=.1$ & 600 \\

    Reputation scope & NoRep; Shared-1; Style-6; Local-12 &
    33.3\% & 33.3\% & 33.3\% & 0, 1, 6, or 12 histories &
    $N=6,K=3,\rho=.1$ & 1,200 \\

    Catalog overlap & $\rho\in\{0.1,1.0\}$ &
    33.3\% & 33.3\% & 33.3\% & None &
    $N=6,K=3$ & 600 \\
    \bottomrule
  \end{tabular}%
  }
\end{table}

\subsection{Analytical Benchmark for Controlled Access}
\label{app:controlled-access}

The controlled-inclusion rule makes the upstream Access difference analytically
transparent. Let $F$ denote target presence produced specifically by forced
inclusion,
and let $M$ be the number of platform catalogs that carry the held-out target.
Conditional on $M=m$, under Platform-Centric the user first enters one of the
$N$ platforms uniformly, so
\begin{equation}
  \Pr(F_{\mathrm{PC}}\mid M=m)
  = \frac{m}{N}p_{\mathrm{hit}}.
\end{equation}
User agent mediation queries every platform, and $F$ occurs if at least one of the $m$
target-carrying platforms triggers forced inclusion, giving
\begin{equation}
  \Pr(F_{\mathrm{user\,agent}}\mid M=m)
  = 1-(1-p_{\mathrm{hit}})^m.
\end{equation}
For $1\leq m\leq N$,
\begin{equation}
  1-(1-p_{\mathrm{hit}})^m
  \geq p_{\mathrm{hit}}
  \geq \frac{m}{N}p_{\mathrm{hit}},
\end{equation}
so cross-platform querying weakly increases the probability of controlled
target inclusion relative to platform precommitment.

Under our catalog construction, every item has one primary platform and is
copied independently to each of the remaining platforms with probability
$\rho$. Hence $M=1+\operatorname{Binomial}(N-1,\rho)$, and averaging over
catalog assignment yields
\begin{align}
  \Pr(F_{\mathrm{PC}})
  &= p_{\mathrm{hit}}\frac{1+(N-1)\rho}{N},\\
  \Pr(F_{\mathrm{user\,agent}})
  &= 1-(1-p_{\mathrm{hit}})(1-\rho p_{\mathrm{hit}})^{N-1}.
\end{align}
At the default $N=6$, $\rho=0.1$, and $p_{\mathrm{hit}}=0.8$, these quantities
are $20.0\%$ and $86.8\%$, respectively. They describe the forced-inclusion
component rather than overall Target Present: when forced inclusion is not
triggered, the retriever
may still recover the target in either condition. The calculation therefore
serves as a protocol-level sanity check for Access; the empirical question is
how the resulting opportunities survive shortlisting, ranking, and purchase.

\subsection{Uncertainty Estimation}
\label{app:statistical-analysis}

For contrasts among stateless NoRep conditions, we resample paired queries
separately within each domain. For every contrast involving LocalRep, an
outcome can depend on earlier updates for the same simulated user. We therefore
resample the 12 simulated users within each domain, preserving each user's
complete ordered history and using the same resampled user in paired treatment
arms. Each draw first computes a domain estimate and then takes an unweighted
mean across the three domains. Figures and text report percentile intervals from
50,000 draws. As noted in Section~\ref{sec:exp-setup}, primary intervals
condition on the fixed model, seed, catalog assignment, and observed episode
ordering.

\subsection{Robustness to Catalog Assignment and Evaluation Model}
\label{app:robustness}

The main accountability comparison holds the platform catalogs and generative
model fixed. We conduct two complementary robustness tests to determine whether
the observed attention reallocation depends on either choice. Both tests use
the equal three-policy market, in which Strict, Balanced, and Exaggeration each
control one third of platforms, and compare NoRep with Local-12 using 600
matched requests per treatment cell.

\paragraph{Catalog reassignment.}
Presentation policies remain attached to platform identifiers, while inventories
are reassigned cyclically. In the one-step rotation R1, platform $P_j$ receives
the inventory originally assigned to $P_{j+1}$; in the two-step rotation R2, it
receives the inventory of $P_{j+2}$, with indices taken modulo six. These
rotations preserve every catalog and the aggregate item supply while changing
which inventory is paired with each fixed presentation policy. We repeat both
rotations in all three domains and report unweighted macro estimates.

\paragraph{Evaluation-model sensitivity.}
On Sports \& Outdoors, we keep the matched requests, retrieved candidates,
query generation, and explanation generator fixed. We compare four evaluation-side
models: DeepSeek-V4-Flash, DeepSeek-V4-Pro, GLM-5.2, and
MiniMax-M2.5. The evaluation-side model produces the user agent-ranking and user-decision
judgments and, under LocalRep, the feedback records used in later rounds.

\begin{figure}[t]
  \centering
  \includegraphics[width=\textwidth]{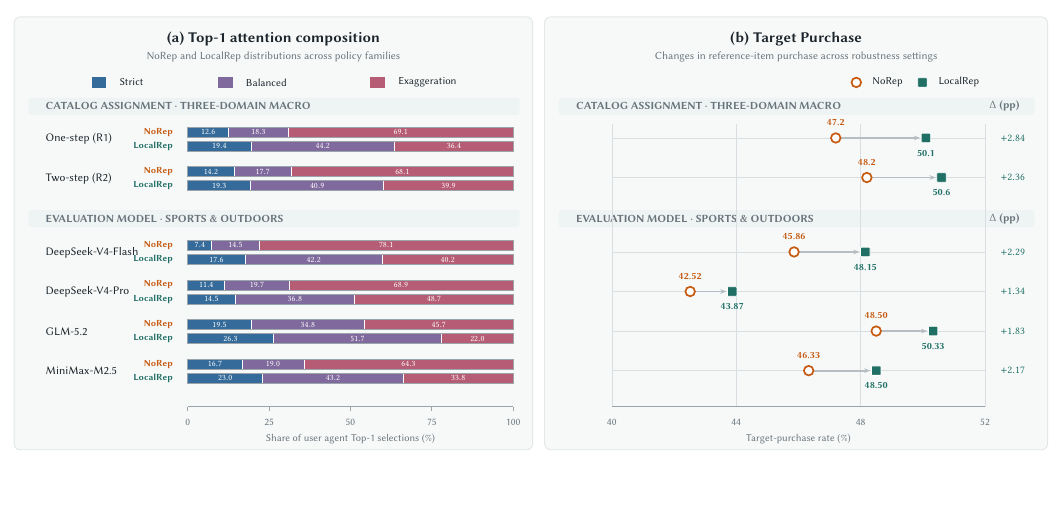}
  \Description{The left panel uses paired 100-percent stacked bars to show how
  Top-1 attention is distributed among Strict, Balanced, and Exaggeration
  platforms under NoRep and LocalRep. The right panel uses paired markers and
  arrows to compare Target Purchase. Rows cover two cyclic catalog rotations
  and four evaluation models.}
  \captionsetup{aboveskip=4pt,belowskip=-8pt}
  \caption{Robustness of the LocalRep accountability result. The left panel
  shows the complete composition of user agent Top-1 attention across the three
  equal-supply policy families under NoRep and LocalRep; the right compares
  Target Purchase, with orange circles and green squares marking the two
  conditions. Catalog-assignment rows are unweighted means across the three
  domains; model-sensitivity rows use Sports \& Outdoors. Purchase differences
  are LocalRep minus NoRep.}
  \label{fig:robustness-summary}
\end{figure}

\begin{table}[t]
  \caption{Catalog-assignment and evaluation-model robustness. Attention
  columns report Exaggeration's share of all user agent Top-1 selections; purchase
  columns report the target-purchase rate. Catalog rotations are unweighted
  three-domain macro estimates, while model comparisons use Sports \&
  Outdoors. All entries are percentages and differences are
  LocalRep minus NoRep.}
  \label{tab:robustness-summary}
  \small
  \setlength{\tabcolsep}{4.1pt}
  \renewcommand{\arraystretch}{1.10}
  \resizebox{\textwidth}{!}{%
  \begin{tabular}{@{}llcrrrcrrr@{}}
    \toprule
    & & & \multicolumn{3}{c}{\textbf{Exaggeration Top-1 share}} &&
      \multicolumn{3}{c}{\textbf{Target Purchase}} \\
    \cmidrule(lr){4-6}\cmidrule(lr){8-10}
    \textbf{Robustness dimension} & \textbf{Setting} & \textbf{Scope} &
    \textbf{NoRep} & \textbf{LocalRep} & \textbf{$\Delta$} &&
    \textbf{NoRep} & \textbf{LocalRep} & \textbf{$\Delta$} \\
    \midrule
    Catalog assignment & One-step rotation (R1) & 3-domain macro &
    69.1 & 36.4 & $-32.7$ && 47.2 & 50.1 & $+2.84$ \\
    Catalog assignment & Two-step rotation (R2) & 3-domain macro &
    68.1 & 39.9 & $-28.2$ && 48.2 & 50.6 & $+2.36$ \\
    \midrule
    Evaluation model & DeepSeek-V4-Flash & Sports &
    78.1 & 40.2 & $-37.9$ && 45.86 & 48.15 & $+2.29$ \\
    Evaluation model & DeepSeek-V4-Pro & Sports &
    68.9 & 48.7 & $-20.2$ && 42.52 & 43.87 & $+1.34$ \\
    Evaluation model & GLM-5.2 & Sports &
    45.7 & 22.0 & $-23.7$ && 48.50 & 50.33 & $+1.83$ \\
    Evaluation model & MiniMax-M2.5 & Sports &
    64.3 & 33.8 & $-30.5$ && 46.33 & 48.50 & $+2.17$ \\
    \bottomrule
  \end{tabular}%
  }
\end{table}

Figure~\ref{fig:robustness-summary} shows the full redistribution of Top-1
attention alongside changes in downstream purchase outcomes. Under R1 and
R2, the largest policy-family share falls from 69.1\% to 44.2\% and from
68.1\% to 40.9\%, respectively. The corresponding Target Purchase increases
are 2.84 points (95\% CI $[0.17,5.55]$) and 2.36 points
(95\% CI $[0.90,3.82]$). Thus, the main accountability result is not tied to
the initial pairing between inventories and presentation policies.

The four evaluation models also show substantial changes in the composition of
Top-1 attention across all three policy families. Target Purchase increases
under every model. Together, these tests show that the accountability pattern
is not specific to one catalog assignment or evaluation model: LocalRep
consistently redistributes Top-1 attention across policy families, while Target
Purchase moves in the same direction across settings.

\subsection{How Does Reputation-Memory Scope Affect the Market?}
\label{app:rep-scope}

The user-side architecture studied in this paper does not assume a central
reputation service. Each simulated user's user agent can retain its own outcome history,
but sharing that history across users would require an additional aggregation
and trust layer. We therefore use shared reputation as a counterfactual to test
whether fragmented local evidence materially limits accountability.
We compare \emph{Shared-1}, which pools all six user profiles into one platform
history; \emph{Style-6}, which maintains one history per profile; and the
default \emph{Local-12}, which uses two simulated users per profile. Every
user's user agent retains at most $W=10$ claim-reliability records per platform.

Each simulated user receives an ordered sequence of requests, allowing that
user's user agent history to develop across interactions. This represents simulated
continuity rather than reconstructing longitudinal Amazon customers. The
comparison below therefore concerns the scope and cold-start behavior of
simulated reputation evidence.

\begin{figure}[t]
  \centering
  \includegraphics[width=\textwidth]{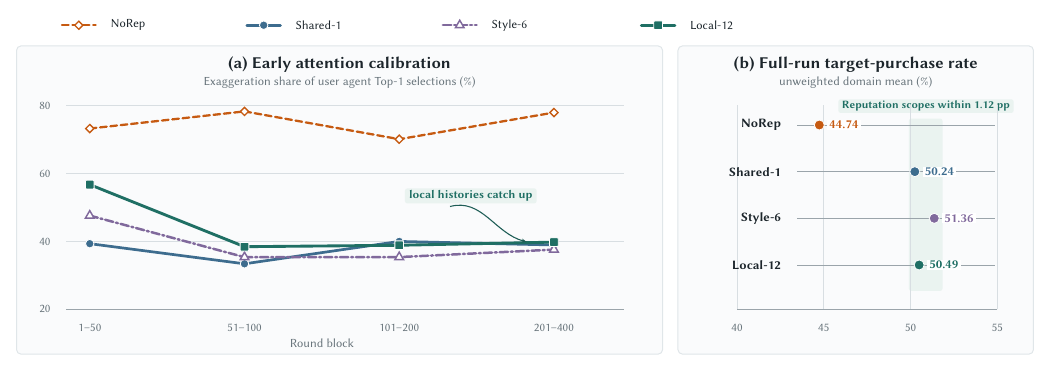}
  \Description{The left panel shows Exaggeration's share of user agent Top-1 selections in
  four early round blocks. Shared reputation reduces Exaggeration attention fastest
  in the first block, but the three reputation variants converge after local
  histories begin to accumulate. The right panel shows that their full-run
  target-purchase rates are similar.}
  \captionsetup{aboveskip=4pt,belowskip=0pt}
  \caption{Reputation scope in the equal three-policy market. Left: unweighted
  domain means for Exaggeration's share of all user agent Top-1 selections over early round
  blocks. Right: full-run macro target-purchase rates. Shared-1 pools feedback
  across profiles, Style-6 keeps one platform history per profile, and Local-12
  uses two independent simulated users per profile. Sharing accelerates the initial
  attention response but does not improve aggregate purchase beyond local
  histories.}
  \label{fig:reputation-scope}
\end{figure}

Figure~\ref{fig:reputation-scope} shows that sharing primarily changes the
speed of initial calibration. In rounds 1--50, Exaggeration receives 39.3\% of
Top-1 attention under Shared-1, compared with 47.6\% under Style-6, 56.7\%
under Local-12, and 73.2\% under NoRep. Pooled feedback therefore
allows the user agent to recognize a persistent presentation pattern sooner. The
advantage is short-lived: by rounds 101--200, the three reputation scopes differ
in Exaggeration attention by at most 4.6 pp, and by rounds 201--400 they differ
by at most 2.2 pp. Local histories need not fill every slot before their ranking effect
stabilizes; a small amount of repeated platform evidence is sufficient to
calibrate confidence in the fixed policies tested here.

Faster pooling also does not produce a corresponding purchase advantage. The
full-run macro target-purchase rates are 50.24\% for Shared-1, 51.36\% for
Style-6, and 50.49\% for Local-12. Reputation scope changes how quickly
platform-level evidence becomes available, whereas purchase still depends on
the current candidate set, item fit, shortlist composition, and the user's
final choice. Thus Shared-1 is best interpreted as a cold-start control:
within this fixed-policy testbed, cross-user pooling accelerates early
attention calibration, while partitioned memories converge to a similar
aggregate pattern after evidence accumulates.

\subsection{What the Reputation Memory Learns}
\label{app:reputation-labels}

\begin{figure}[t]
  \centering
  \includegraphics[width=\columnwidth]{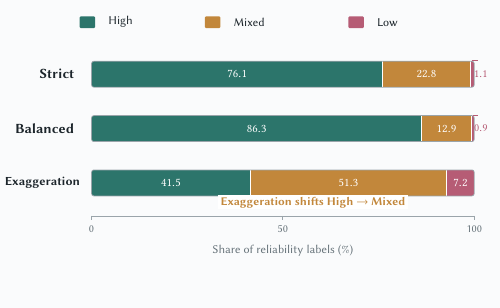}
  \Description{Three horizontal stacked bars show High, Mixed, and Low
  reliability labels for Strict, Balanced, and Exaggeration platforms. Strict and
  Balanced are mostly High, while Exaggeration shifts much of the mass from High to
  Mixed; Low remains a small minority.}
  \captionsetup{aboveskip=4pt,belowskip=-8pt}
  \caption{Claim-reliability labels among purchase-grounded Local-12 updates,
  pooled across domains. Percentages condition on the selected platform's
  presentation-policy family. Reputation primarily distinguishes calibrated claims
  from partially overstated ones rather than blacklisting platforms.}
  \label{fig:reputation-labels}
\end{figure}

The reputation updater records claim calibration rather than a generic product
rating. Figure~\ref{fig:reputation-labels} audits the purchase-grounded memory
entries generated under the default Local-12 equal three-policy market. Strict
and Balanced purchases produce entries labeled High in 76.1\% and 86.3\% of
updates, respectively. Exaggeration produces only 41.5\% High entries; its largest
category is instead Mixed at 51.3\%. Outcome feedback therefore does not
operate mainly by assigning categorical penalties. It accumulates graded
evidence that an otherwise plausible explanation repeatedly expressed greater fit or
certainty than the later experience supported, giving the user agent a basis for
tempering similar claims in future rounds.

The Low category is uncommon because Exaggeration may overstate fit or downplay
trade-offs but is prohibited from fabricating objective product attributes. The plotted
distribution is conditional on a candidate item being purchased and consequently
producing a reputation update; it is an audit of the evidence entering memory,
not an unconditional estimate of each policy family's recommendation quality.

\subsection{Accountability Case Studies}
\label{app:accountability-cases}

Table~\ref{tab:accountability-cases} illustrates two complementary behaviors
of the update loop. The excerpts lightly abridge generated text and anonymize
platform identifiers; the observed reliability labels, ranks, and purchase
outcomes are unchanged.

\begin{table}[t]
  \caption{Selected accountability cases from Musical Instruments. The first
  separates claim reliability from satisfaction. The second shows how several
  moderate expectation gaps accumulate into a later ranking adjustment.}
  \label{tab:accountability-cases}
  \small
  \setlength{\tabcolsep}{5pt}
  \renewcommand{\arraystretch}{1.13}
  \begin{tabular}{@{}p{0.17\textwidth}p{0.49\textwidth}p{0.25\textwidth}@{}}
    \toprule
    \textbf{Mechanism} & \textbf{Observed evidence chain} &
    \textbf{Reputation consequence} \\
    \midrule
    Candid limitation is not punished &
    A user requested a 10-inch guitar speaker that stayed clean while handling
    crunch. Platform A recommended a high-power 10-inch speaker but explicitly
    disclosed that it was a 16-ohm bass speaker whose response could differ
    from a guitar speaker. After purchase, the user reported that clean volume
    was good but the speaker lacked the desired crunch and midrange bite, and
    returned it. &
    \textbf{High.} The later experience confirmed the exact limitation that
    the recommendation explanation had disclosed. The mechanism evaluates expectation calibration,
    not whether the user was ultimately satisfied. \\

    Repeated partial overstatement tempers later confidence &
    Platform B accumulated Mixed entries after three purchases: a bass case
    was sturdy and fit correctly but was heavier than promised; a braided cable
    worked but was stiffer than suggested; and a quick-change capo functioned
    but its clamp was less secure than the ``hassle-free'' framing implied. In
    a later tuner request, Platform B again used a highly confident recommendation explanation
    (``perfect,'' ``reliable,'' and ``effortless''). The user agent retained the candidate
    item but cited the Mixed history, discounted the overconfident wording, and
ranked it second. A candid recommendation from a platform with a High-reliability
    history was ranked first and purchased. &
    \textbf{Accumulated Mixed evidence.} LocalRep does not blacklist a platform
    after one disappointing outcome. Repeated, partially supported
    overstatement gradually lowers confidence in later claims while leaving the
    current item eligible to compete. \\
    \bottomrule
  \end{tabular}
\end{table}

\subsection{When Catalog Overlap Turns Item Competition into Explanation Competition}
\label{app:overlap-persuasion}

The signaling results in the main text compare markets whose platforms usually
submit different items. To isolate the boundary at which presentation becomes
the primary competitive margin, we compare the default overlap $\rho=0.1$ with
complete overlap $\rho=1.0$ under NoRep and the equal 33.3\%--33.3\%--33.3\%
policy composition. We define a \emph{direct same-item policy comparison} as
an episode in which the item ranked Top-1 by the user agent was simultaneously
submitted by at least two presentation-policy families.

\begin{figure}[t]
  \centering
  \includegraphics[width=\columnwidth]{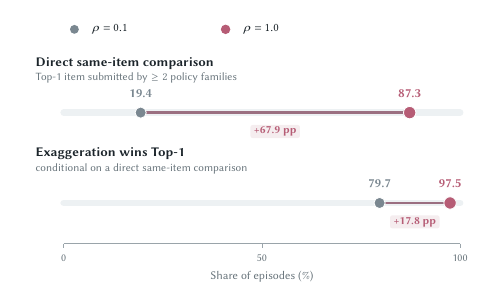}
  \Description{Two horizontal comparisons contrast default and complete
  catalog overlap. The prevalence of direct same-item policy comparisons rises
  from about nineteen to eighty-seven percent, and Exaggeration's conditional
  Top-1 win rate rises from about eighty to ninety-eight percent.}
  \captionsetup{aboveskip=4pt,belowskip=-8pt}
  \caption{Presentation competition under NoRep. A direct same-item comparison
  occurs when the user agent's Top-1 item was submitted by at least two policy families.
  With complete overlap, these comparisons become the norm and Exaggeration wins
  nearly all of them despite controlling 33.3\% of platform supply. Values are
  unweighted domain means.}
  \label{fig:overlap-persuasion}
\end{figure}

At $\rho=0.1$, such direct comparisons occur in 19.4\% of episodes, and
Exaggeration wins 79.7\% of them. At $\rho=1.0$, the same Top-1 item is presented
under multiple policies in 87.3\% of episodes, and Exaggeration wins 97.5\% of
those comparisons (Figure~\ref{fig:overlap-persuasion}). As inventories converge,
item-selection differences recede and the user agent increasingly chooses among
alternative presentations of the same item. The persuasive advantage is
strongest precisely where competition is least differentiated by the
underlying product. Catalog convergence therefore shifts the competitive
margin toward how the same item is presented, making the user agent's treatment of
platform claims increasingly consequential for attention allocation.

%% file: sections/appendix.tex
\section{Style Operationalization and Prompt Design}
\label{app:style-design}

This appendix documents the controlled styles used to introduce heterogeneity
into user requests and platform recommendation explanations. We use \emph{style} as an
operational term: the labels specify prompt-level behavioral patterns rather
than psychometric personality constructs or exhaustive population categories.
Their design draws on recurring distinctions in search behavior, consumer
choice, recommendation explanation, and persuasion reviewed below.

The current design contains two style families. A user profile governs how a
need is expressed and how the resulting purchase is privately evaluated. A
platform presentation policy governs how a retrieved item is presented.

\subsection{User Request and Evaluation Profiles}
\label{app:user-profiles}

The six profiles draw on four recurring dimensions of search and consumer
choice. First, search-goal taxonomies distinguish transactional from
informational and other goal-directed activity
\cite{broder2002taxonomy,rose2004understanding}. Second, preferences can be
constructed around task-dependent trade-offs \cite{bettman1998constructive} or
elicited through explicit, often attribute-level critiques
\cite{chen2012critiquing}. Third, domain expertise changes query vocabulary,
resource use, and search behavior \cite{white2009expertise}. Finally,
exploratory search emphasizes open-ended learning and discovery
\cite{marchionini2006exploratory}, while consumer research distinguishes
utilitarian from hedonic shopping value \cite{babin1994work}. We translate
these dimensions into the prompt-level profiles in
Table~\ref{tab:user-profiles}.

\begin{table}[t]
  \caption{Operational user profiles. The request and evaluation columns
  summarize the query-style guidance and review-specific rubrics; final
  decisions use the profile guidance together with common decision rules.}
  \label{tab:user-profiles}
  \small
  \setlength{\tabcolsep}{4pt}
  \renewcommand{\arraystretch}{1.10}
  \begin{tabular}{@{}p{0.18\textwidth}p{0.22\textwidth}p{0.34\textwidth}p{0.15\textwidth}@{}}
    \toprule
    \textbf{Profile} & \textbf{Request realization} &
    \textbf{Private evaluation lens} & \textbf{Primary grounding} \\
    \midrule
    \texttt{transactional\_\allowbreak direct} &
    A short, direct, purchase-oriented request with a clear immediate goal and
    little ancillary context. &
    Prioritizes whether the item solves the stated buying goal with minimal
    friction; feedback is concise and practical, with a direct reaction to a
    miss on the immediate need. &
    Transactional search goals
    \cite{broder2002taxonomy,rose2004understanding}. \\

    \texttt{informational\_\allowbreak use\_\allowbreak case} &
    Describes a concrete activity, context, or desired outcome and leaves the
    system to infer an appropriate product type. &
    Judges fit to the concrete use case rather than general product quality;
    partial fit can yield a measured or mixed review that names the specific
    match or mismatch. &
    Goal-oriented information seeking
    \cite{rose2004understanding}. \\

    \texttt{comparative\_\allowbreak evaluation} &
    States relevant trade-offs---for example, durability versus portability or
    simplicity versus advanced control---without requiring a preselected
    pairwise comparison. &
    Evaluates the item against the expressed dimensions and identifies the main
    compromise, trade-off, or standout advantage. &
    Constructive, task-contingent choice
    \cite{bettman1998constructive}. \\

    \texttt{attribute\_\allowbreak constraint} &
    Emphasizes hard requirements such as compatibility, interface, size,
    material, format, installation limits, budget range, or required features. &
    Treats violation of a stated hard constraint as decision-critical even when
    the item has other attractive qualities. &
    Attribute-level preference articulation
    \cite{chen2012critiquing}. \\

    \texttt{novice\_\allowbreak advice} &
    Uses nontechnical language and asks for an easy-to-use, low-regret option or
    guidance about which considerations matter. &
    Is relatively forgiving when an item is usable, but reacts negatively when
    the explanation made ease, safety, or beginner suitability sound stronger than
    the experienced outcome. &
    Expertise-sensitive search behavior
    \cite{white2009expertise}. \\

    \texttt{exploratory\_\allowbreak hedonic} &
    Welcomes discovery, adjacent products, bundles, or unexpected substitutes
    and may express desired aesthetics, feel, identity, or emotional experience. &
    Prioritizes experiential fit, enjoyment, aesthetics, and discovery value;
    practical imperfections can be tolerated when the experience is satisfying. &
    Exploratory search and hedonic value
    \cite{marchionini2006exploratory,babin1994work}. \\
    \bottomrule
  \end{tabular}
\end{table}

These profiles are designed for scenario coverage rather than factor
identification. In particular, \texttt{exploratory\_hedonic} deliberately
combines an open-ended search mode with an experiential value orientation, and
\texttt{novice\_advice} combines low domain expertise with an operational
preference for easy, low-regret choices. We therefore do not interpret
between-profile differences as causal effects of a single psychological
construct.

\paragraph{Cross-stage consistency and privacy.}
Each simulated user has a private profile that is used consistently to generate
requests, make final decisions, and, when applicable, write post-purchase
reviews. The profile is internal to the user simulation: neither platforms nor the user agent
receive its label.

The held-out positive interaction is likewise private. Its title, rating, and
original review can serve as a preference anchor when a post-purchase review is
generated, but the target is unavailable to platforms, the user agent, and the
final-decision model,
and the review prompt prohibits identifying it.

In the default LocalRep setup, we instantiate two simulated users per profile,
for 12 users in total. Each user has their own user agent, so platform histories are
user-specific and persist only across that user's interactions. These
simulated continuities do not reconstruct longitudinal Amazon customers.

\subsection{Platform Presentation Policies}
\label{app:platform-policies}

Platform presentation policies are platform-side communication policies, not platform
personalities. Recommendation explanations can serve goals including
transparency, decision support, trust, and persuasion, although these goals can
conflict
\cite{tintarev2012evaluating}; two-sided messages additionally show that
disclosing unfavorable information can alter both source credibility and
persuasive impact depending on how the message is constructed
\cite{eisend2006twosided}. The Persuasion Knowledge Model provides the broader
motivation for recipients to reason about a seller's goals and tactics
\cite{friestad1994persuasion}. This distinction is increasingly consequential
for LLM-mediated recommendation: generated explanations can remain persuasive
while incorporating insufficiently credible information
\cite{qin2024beyond}, persuasive wording can
change users' choices \cite{rahman2026persuasive}, and seller agents can
strategically adapt product presentations for AI shoppers
\cite{allouah2025agentbuying}.

We operationalize this communication spectrum with the three policies in
Table~\ref{tab:platform-policies}. The middle policy is an experimental reference
condition, not a category claimed by prior theory. All policies share the same
prohibition on inventing objective product attributes. This narrow boundary is
motivated by the longstanding advertising-substantiation principle that express
and implied objective claims should have a reasonable basis
\cite{ftc1984substantiation}. We use the label \emph{Exaggeration} in the main
text because the policy deliberately increases
confidence and selectively downplays limitations. Its implementation identifier
is \texttt{promotional\_exaggeration}, emphasizing that the intervention changes
promotional framing without permitting invented objective attributes.

\begin{table}[t]
  \caption{Operational platform presentation policies. Each policy generates a one- to
  three-sentence recommendation explanation after candidate retrieval.}
  \label{tab:platform-policies}
  \small
  \setlength{\tabcolsep}{5pt}
  \renewcommand{\arraystretch}{1.10}
  \begin{tabular}{@{}p{0.20\textwidth}p{0.33\textwidth}p{0.37\textwidth}@{}}
    \toprule
    \textbf{Policy} & \textbf{Framing instruction} &
    \textbf{Disclosure and evidential boundary} \\
    \midrule
    Strict (\texttt{strict\_\allowbreak transparent}) &
    Use an evidence-first, non-sales-oriented presentation and reflect the item
    as accurately as the visible data permit. &
    Explicitly state relevant limitations, missing evidence, uncertainty,
    partial fit, or trade-offs. Avoid hype, unsupported superiority, and
    overconfidence; never invent hard facts or objective attributes. \\

    Balanced (\texttt{balanced\_\allowbreak advisor}) &
    Present the strongest supported benefit for the user's need in a neutral,
    concise, and practically helpful tone. &
    Mention an important limitation or trade-off when relevant. Remain neither
    overly critical nor promotional, and never invent hard facts or objective
    attributes. \\

    Exaggeration (\texttt{promotional\_\allowbreak exaggeration}) &
    Use confident, enthusiastic, selectively positive framing; emphasize
    convenience and positive outcomes, and permit subjective superlatives. &
    The explanation may downplay trade-offs and make a partial or adjacent fit sound
    more certain, but it may not assert absent specifications, compatibility,
    variants, effects, or other objective attributes. \\
    \bottomrule
  \end{tabular}
\end{table}

\subsection{Experimental Controls and Information Boundaries}
\label{app:style-controls}

The presentation policy is applied only after each platform retrieves its candidate,
so it cannot change inventory, retrieval score, or the submitted item. All policies
receive the same structured need and item-evidence schema, share a one- to
three-sentence budget, and follow the same prohibition against inventing
objective attributes, compatibility, variants, or effects. The manipulated dimensions
are therefore framing, disclosure, and claim calibration.

To construct its recommendation explanation, the platform sees the item title,
category, and description; the user agent ranks using the title, category, explanation,
and, when enabled, the platform's
local history. The held-out target and platform policy label are absent from the
user agent ranking and final-decision inputs. After a purchase, the reputation updater
compares the recommendation explanation with the simulated review without receiving
either field. Complete
payloads and role prompts appear in Appendix~\ref{app:prompts}.

Presentation policies remain attached to platform identifiers across rounds so that
outcome-grounded histories refer to a stable source. Homogeneous NoRep cells compare the
three policies under the same base requests, while mixed cells study how they
compete with and without outcome histories. Policy labels are retained only in
the experiment log for analysis.

\section{Selected Experimental Prompt Templates}
\label{app:prompts}

This appendix presents the prompt components that define the main experimental
roles and treatments. Bracketed terms mark dynamic inputs, while repeated
implementation scaffolding is omitted. In the prompt cards, \emph{Rep} includes
platform history and \emph{NoRep} does not; profile-specific instructions follow
Table~\ref{tab:user-profiles}.

\subsection{Held-Out-Interaction-Conditioned Query Generation}
\label{app:prompt-query}

Among pre-purchase stages, only query generation observes the held-out final
interaction. It uses the target to construct a plausible request while
prohibiting a unique or post hoc item description. Neither the user agent ranker nor the
final-decision model receives the target; its private post-purchase use is documented in
Appendix~\ref{app:prompt-accountability}.

\begin{promptbox}{UserPrompt}{UserPromptBG}{QUERY GENERATION}
\PromptField{SYSTEM}{You generate realistic e-commerce user search intent
records in strict JSON. Never return markdown.}

\PromptField{DYNAMIC INPUT}{\texttt{[product domain]},
\texttt{[profile-specific style instruction]},
\texttt{[older interactions]}, and the held-out item's
\texttt{[title]}, \texttt{[rating]}, and \texttt{[review]}.}

\PromptField{TASK}{Infer user preference from history. Generate one plausible
pre-purchase query for which the intended target item would be a plausible
strong answer while several credible alternatives remain. Follow the required
style.}

\PromptRule{Express the user's underlying intent, use case, taste, or broad
constraints; prefer natural goals and soft preferences over exact target
identifiers.}
\PromptRule{Infer the likely shopping situation and wording from the target
title and review rather than copying the title into the query.}
\PromptRule{Do not introduce hard requirements that would rule out the target,
or turn a review-only detail into a ``must,'' ``need,'' ``required,'' or exact
compatibility requirement.}
\PromptRule{Avoid both a uniquely identifying query and one so broad that many
unrelated products would be equally good answers.}
\PromptRule{Do not mention internal fields such as
\texttt{parent\_asin}, \texttt{user\_id}, or \texttt{timestamp}.}

\PromptSchema{Output: \{"query\_text": "[one natural shopping query]"\}}
\end{promptbox}

\subsection{Need Structuring by the user agent}
\label{app:prompt-structuring}

Before retrieval, the user agent converts the reconstructed request and observable
history into the two-part representation used by every platform. This is a
separate model call from query generation: the structured representation is not shown to
the final user, and it contains no target identity.

\begin{promptbox}{UserAgentPrompt}{UserAgentPromptBG}{user agent NEED STRUCTURING}
\PromptField{SYSTEM}{You are a strict JSON generator for recommendation query
understanding.}

\PromptField{DYNAMIC INPUT}{\texttt{[older user history]} and the current
\texttt{[query\_text]}.}

\PromptField{TASK}{Convert the query into a compact JSON object for downstream
retrieval. Infer longer-term preference from history while keeping current
intent tightly grounded in the current query.}

\PromptRule{\texttt{long\_term\_need} is at most 30 words and reflects the
broader, more stable need or preference pattern, inferred mainly from history.}
\PromptRule{\texttt{current\_need} is at most 25 words, rewrites the current
query for retrieval, preserves its meaning, and adds no brands, models, or
specifications not implied by that query.}
\PromptRule{When history is informative, do not make
\texttt{long\_term\_need} merely a paraphrase of the current query; keep
\texttt{current\_need} close to the immediate task.}

\PromptSchema{Output: \{"long\_term\_need":"[at most 30 words]",
"current\_need":"[at most 25 words]"\}}
\end{promptbox}

\subsection{Platform Explanation Treatment}
\label{app:prompt-explanation}

Each platform receives the same structured need and the evidence visible for its
retrieved item. The following shared wrapper fixes the factual boundary;
the orange policy block is the treatment-specific insertion and retains the
operative distinctions among the three policies in full. The final variant is
the Exaggeration condition named in the main text and retains the prompt wording
used in the experiment.

The prompt cards reproduce implementation field names such as
\texttt{pitch}, \texttt{bidding\_pitch}, and \texttt{merchant\_pitch}; in the
surrounding paper, these messages are referred to as recommendation explanations.

\begin{promptbox}{PlatformPrompt}{PlatformPromptBG}{PLATFORM RECOMMENDATION EXPLANATION}
\PromptField{SYSTEM}{You write product recommendation explanations for a
participating platform.}

\PromptField{DYNAMIC INPUT}{\texttt{[platform ID]},
\texttt{[explanation policy]}, \texttt{[policy guidance]},
\texttt{[structured user need]}, and \texttt{[candidate item evidence]}.}

\PromptField{TASK}{Write a concise recommendation explanation for the candidate item.}

\PromptRule{Write one to three sentences and follow the specified policy;
do not switch tone or objective.}
\PromptRule{Return plain text only.}

\PromptVariant{STRICT TRANSPARENT}{Use a strict, transparent, evidence-first
style. Reflect the item as accurately as possible rather than trying to sell it.
Ground claims in the item data, and explicitly mention relevant limitations,
missing evidence, uncertainties, or trade-offs when they matter for the user's
need. If the evidence is weak, incomplete, low-rated, or only a partial match,
say so clearly. Avoid hype, persuasion, unsupported superiority, and
overconfidence; do not invent hard facts or objective attributes.}

\PromptVariant{BALANCED ADVISOR}{Use a neutral, practical recommendation style.
Explain the strongest user-relevant benefit supported by the item data, and
briefly mention an important trade-off or limitation when relevant. Be helpful
and concise, like a fair shopping advisor: neither overly critical nor overly
promotional. Do not invent hard facts or objective attributes.}

\PromptVariant{PROMOTIONAL EXAGGERATION}{Use promotional exaggeration:
confident, enthusiastic, selectively framed, and strongly persuasive. Frame the
item as an especially strong fit, convert partial evidence into confident fit
language, and emphasize convenience and positive outcomes. Present an adjacent substitute as a smart
practical solution and minimize category, compatibility, or evidence gaps.
Subjective superlatives, downplayed trade-offs, and greater certainty than the
evidence warrants are permitted, but inventing hard facts or objective
attributes is not.}

\PromptSchema{Output: [one plain-text recommendation explanation of 1--3 sentences]}
\end{promptbox}

\subsection{Shortlisting by the user agent: Rep versus NoRep}
\label{app:prompt-shortlist}

Candidate order is shuffled before this call. In both conditions the user agent sees
the explicit request, its structured version, visible item fields, and the
current recommendation explanation. Only the Rep variant receives
\nolinkurl{history_reputation_profile}; neither condition receives the platform policy
label. Both conditions use the same current-round ranking instructions; the Rep
condition receives only the additional history-dependent guidance shown below.

\begin{promptbox}{UserAgentPrompt}{UserAgentPromptBG}{user agent SHORTLIST}
\PromptField{SYSTEM}{You are a strict ranking judge for recommendation
platforms.}

\PromptField{ROLE}{Prioritize user benefit. Choose exactly
\texttt{[K]} of \texttt{[N]} platform candidates and rank them from best to
worst. You cannot decide whether the user buys.}

\PromptField{COMMON EVIDENCE}{\texttt{[explicit query]},
\texttt{[structured understanding]}, and a shuffled candidate list containing
\texttt{platform\_id}, \texttt{candidate\_item}, and
\texttt{bidding\_pitch}.}

\PromptRule{Make the ranking judgment from the current information available
to the user agent.}
\PromptRule{Rank by apparent user benefit using the explicit query, structured
understanding, visible item fields, and current recommendation explanation.}
\PromptRule{Evaluate the current recommendation explanation together with visible item
information and the user's goal. Explain the recommendation in
\texttt{user\_agent\_reason} from the current item evidence and recommendation explanation.}

\PromptVariant{REP-ONLY ADDITION}{Historical reputation profiles are available.
With no history, evaluate the current recommendation explanation together with visible item
information and the user's goal, and explain the recommendation from current
item evidence and the recommendation explanation. Treat reputation as a claim-calibration tool,
never as a platform score: High reliability is not a ranking bonus, and Low or
Mixed reliability is not a ranking penalty by itself. Interpret
\texttt{claim\_reliability} as claim--evidence consistency: High means past
claims were supported by post-purchase evidence or accurately caveated; Mixed
means a relevant limitation or uncertainty was omitted or understated; Low
means a core claim was contradicted by post-purchase evidence or was materially
misleading. Use relevant records to judge the reliability of specific claims in
the current recommendation explanation. Separate value supported by visible evidence from value
asserted mainly through the platform's confidence, emphasis, or wording, then
compare candidates by their evidence-calibrated substantive fit for the user.}

\PromptRule{Return exactly \texttt{[K]} unique platform IDs. Do not add a
no-purchase option; the user decides that later.}

\PromptSchema{Output: \{"ranked\_items":
[\{"platform\_id":"P1", "user\_agent\_reason":"[item-specific reason]"\}, ...]\}}
\end{promptbox}

\subsection{Final Purchase Decision}
\label{app:prompt-decision}

The final user observes the user agent's shortlist rather than the full item-evidence
payload. Its private profile is supplied here to keep request construction and
evaluation consistent, but that label was hidden from the platforms and user agent.

\newpage
\begin{promptbox}{FeedbackPrompt}{FeedbackPromptBG}{FINAL USER DECISION}
\PromptField{SYSTEM}{You simulate final user purchase decisions and return
strict JSON only.}

\PromptField{VISIBLE INPUT}{\texttt{[query]}, private
\texttt{[profile instruction]}, and shortlisted entries containing
\texttt{rank}, \texttt{platform\_id}, \texttt{item\_title},
\texttt{item\_category}, \texttt{merchant\_pitch}, and
\texttt{user\_agent\_reason}.}

\PromptRule{The user may choose any shown item, including a lower-ranked one,
or no purchase. Treat \texttt{user\_agent\_reason} as visible shopping advice and
platform recommendation explanations as descriptions that may shape expectations.}
\PromptRule{Prefer the item that appears most useful and best matched to this user's stated need,
considering rank, recommendation explanation, user agent advice, and profile together. Choose no purchase
only when the shown options do not feel good enough for the profile.}
\PromptRule{Assess the current recommendation explanation and user agent advice together with the
other visible evidence.}

\PromptVariant{REP-ONLY ADDITION}{If \texttt{user\_agent\_reason} includes reputation
information, use it to calibrate confidence in the recommendation explanation.}

\PromptSchema{Output: \{"decision":"purchase|no\_purchase",
"selected\_platform\_id":"P1", "reason":"[brief reason]"\}}
\end{promptbox}

The Platform-Centric condition uses the same private profile and decision space,
but its role prompt states that the user has already entered one uniformly
sampled platform and that there is no user agent, cross-platform shortlist, or historical
reputation signal. It therefore receives only that platform's ranked items and
recommendation explanations. We omit the full condition-specific wrapper because it
adds no
additional decision rule.

\subsection{Post-Purchase Accountability Loop}
\label{app:prompt-accountability}

The two model calls below occur only after a purchase when reputation is enabled.
First, a private user review supplies an outcome signal; second, the user agent-side
updater compares the purchased platform's pre-purchase recommendation explanation
with that review and
writes a bounded claim-reliability record for the purchased platform. Only the
purchased platform is updated.

\begin{promptbox}{FeedbackPrompt}{FeedbackPromptBG}{POST-PURCHASE USER REVIEW}
\PromptField{SYSTEM}{You simulate realistic post-purchase user reviews. Return
plain text only.}

\PromptField{DYNAMIC INPUT}{\texttt{[query]}, private
\texttt{[profile instruction and review rubric]}, hidden
\texttt{[target outcome]}, \texttt{[purchased item evidence]}, and the
\texttt{[pre-purchase recommendation explanation]}.}

\PromptRule{Write one short first-person review of the purchased item only,
in one to three sentences; it may be positive, negative, or mixed.}
\PromptRule{Use the profile-specific rubric as the evaluation lens and the
hidden target as the user's true hoped-for result when interpreting the user's
likely goals. Never mention or identify that target, internal fields, alternate
products, or the experiment.}
\PromptRule{Respond as a real user would after experiencing the purchased item:
treat the recommendation explanation as part of what the user expected; express an expectation gap
when the actual item is thinner, less specific, less convenient, less compatible,
lower quality, or less complete than the query and recommendation explanation reasonably suggested,
while treating clearly disclosed limitations, uncertainties, or trade-offs as
expected rather than surprising unless the experience is worse than disclosed.}
\PromptRule{A usable item may still receive a mixed review when an omitted or
understated limitation leaves the need partly unresolved. Neutral or mixed
reviews are allowed when realistic, but must state the specific reason the item
worked, partly worked, or missed the need. Do not invent specifications,
reviews, or guarantees.}

\PromptSchema{Output: [one plain-text first-person review of 1--3 sentences]}
\end{promptbox}

\newpage
\begin{promptbox}{UserAgentPrompt}{UserAgentPromptBG}{REPUTATION UPDATE}
\PromptField{SYSTEM}{You create concise platform reputation memories from user
reviews.}

\PromptField{OBSERVABLE INPUT}{\texttt{[query + structured need]},
\texttt{[platform + purchased item]}, \texttt{[recommendation explanation + user agent reason]}, and
\texttt{[review]}.}

\PromptRule{Assess whether the recommendation explanation calibrated expectations, overstated fit,
omitted constraints, or acknowledged uncertainty; judge claim reliability, not
overall satisfaction.}
\PromptVariant{HIGH}{The recommendation explanation was candid and calibrated; the review confirms
its main claims, disclosed trade-off, or stated uncertainty. Do not penalize a
disclosed limitation merely because it later occurs.}
\PromptVariant{MIXED}{The recommendation explanation was partly accurate but left a meaningful
limitation, uncertainty, or secondary promise less clear than the later review.}
\PromptVariant{LOW}{The review contradicts the recommendation explanation, or an undisclosed or
understated core limitation causes clear disappointment.}
\PromptField{OUTPUT}{\texttt{claim\_reliability=high|mixed|low;} followed by an
evidence-based \texttt{note} of at most 30 words.}
\end{promptbox}

The resulting record is appended to the purchased platform's reputation history
for that simulated user's user agent and is available to future Rep shortlist
calls. NoRep calls omit the history field entirely rather than receiving an
empty or masked profile.

%% file: references.bib
@article{adomavicius2005toward,
  author  = {Gediminas Adomavicius and Alexander Tuzhilin},
  title   = {Toward the Next Generation of Recommender Systems: A Survey of the State-of-the-Art and Possible Extensions},
  journal = {IEEE Transactions on Knowledge and Data Engineering},
  volume  = {17},
  number  = {6},
  pages   = {734--749},
  year    = {2005},
  month   = jun,
  doi     = {10.1109/TKDE.2005.99}
}

@inproceedings{peng2025survey,
  author    = {Qiyao Peng and Hongtao Liu and Hua Huang and Jian Yang and Qing Yang and Minglai Shao},
  title     = {A Survey on {LLM}-powered Agents for Recommender Systems},
  booktitle = {Findings of the Association for Computational Linguistics: EMNLP 2025},
  address   = {Suzhou, China},
  pages     = {11574--11583},
  year      = {2025},
  publisher = {Association for Computational Linguistics},
  doi       = {10.18653/v1/2025.findings-emnlp.620},
  url       = {https://aclanthology.org/2025.findings-emnlp.620/}
}

@misc{lin2026roadmap,
  author  = {Xinyu Lin and Yashar Deldjoo and Sunhao Dai and Honghui Bao and Xiaopeng Ye and Fatemeh Nazary and Wenjie Wang and Tommaso Di Noia and Jun Xu and Tat-Seng Chua},
  title   = {Autonomous Information Seeking: A Roadmap for Agentic Recommender Systems},
  howpublished = {arXiv preprint arXiv:2607.04433},
  year    = {2026},
  doi     = {10.48550/arXiv.2607.04433},
  url     = {https://arxiv.org/abs/2607.04433}
}

@article{huang2025recommender,
  author    = {Xu Huang and Jianxun Lian and Yuxuan Lei and Jing Yao and Defu Lian and Xing Xie},
  title     = {Recommender {AI} Agent: Integrating Large Language Models for Interactive Recommendations},
  journal   = {ACM Transactions on Information Systems},
  volume    = {43},
  number    = {4},
  articleno = {96},
  numpages  = {33},
  year      = {2025},
  doi       = {10.1145/3731446}
}

@misc{hu2026agenticrs,
  author  = {Jinxin Hu and Hao Deng and Lingyu Mu and Hao Zhang and Shizhun Wang and Yu Zhang and Xiaoyi Zeng},
  title   = {Rethinking Recommendation Paradigms: From Pipelines to Agentic Recommender Systems},
  howpublished = {arXiv preprint arXiv:2603.26100},
  year    = {2026},
  doi     = {10.48550/arXiv.2603.26100},
  url     = {https://arxiv.org/abs/2603.26100}
}

@inproceedings{shang2025agentrecbench,
  author    = {Yu Shang and Peijie Liu and Yuwei Yan and Zijing Wu and Leheng Sheng and Yuanqing Yu and Chumeng Jiang and An Zhang and Fengli Xu and Yu Wang and Min Zhang and Yong Li},
  title     = {{AgentRecBench}: Benchmarking {LLM} Agent-based Personalized Recommender Systems},
  booktitle = {Advances in Neural Information Processing Systems},
  volume    = {38},
  year      = {2025},
  publisher = {Curran Associates, Inc.},
  url       = {https://proceedings.neurips.cc/paper_files/paper/2025/hash/e2d6f7249add096e26679eade1b4cc6f-Abstract-Datasets_and_Benchmarks_Track.html}
}

@article{bichler2026agenticmarkets,
  author    = {Martin Bichler},
  title     = {Agentic Markets},
  journal   = {Electronic Markets},
  volume    = {36},
  articleno = {55},
  year      = {2026},
  doi       = {10.1007/s12525-026-00906-y},
  url       = {https://doi.org/10.1007/s12525-026-00906-y}
}

@inproceedings{zhang2024generative,
  author    = {An Zhang and Yuxin Chen and Leheng Sheng and Xiang Wang and Tat-Seng Chua},
  title     = {On Generative Agents in Recommendation},
  booktitle = {Proceedings of the 47th International ACM SIGIR Conference on Research and Development in Information Retrieval},
  address   = {Washington, DC, USA},
  pages     = {1807--1817},
  year      = {2024},
  publisher = {Association for Computing Machinery},
  doi       = {10.1145/3626772.3657844}
}

@inproceedings{xu2025iagent,
  author    = {Wujiang Xu and Yunxiao Shi and Zujie Liang and Xuying Ning and Kai Mei and Kun Wang and Xi Zhu and Min Xu and Yongfeng Zhang},
  title     = {{iAgent}: {LLM} Agent as a Shield between User and Recommender Systems},
  booktitle = {Findings of the Association for Computational Linguistics: ACL 2025},
  address   = {Vienna, Austria},
  pages     = {18056--18084},
  year      = {2025},
  publisher = {Association for Computational Linguistics},
  doi       = {10.18653/v1/2025.findings-acl.928},
  url       = {https://aclanthology.org/2025.findings-acl.928/}
}

@misc{zhang2026usercentric,
  author  = {Luankang Zhang and Hang Lv and Qiushi Pan and Kefen Wang and Yonghao Huang and Xinrui Miao and Yin Xu and Wei Guo and Yong Liu and Hao Wang and Enhong Chen},
  title   = {The Next Paradigm Is User-Centric Agent, Not Platform-Centric Service},
  howpublished = {arXiv preprint arXiv:2602.15682},
  year    = {2026},
  doi     = {10.48550/arXiv.2602.15682},
  url     = {https://arxiv.org/abs/2602.15682}
}

@misc{liu2026governable,
  author  = {Jiahao Liu and Mingzhe Han and Guanming Liu and Weihang Wang and Dongsheng Li and Hansu Gu and Peng Zhang and Tun Lu and Ning Gu},
  title   = {From Hidden Profiles to Governable Personalization: Recommender Systems in the Age of {LLM} Agents},
  howpublished = {arXiv preprint arXiv:2604.20065},
  year    = {2026},
  doi     = {10.48550/arXiv.2604.20065},
  url     = {https://arxiv.org/abs/2604.20065}
}

@misc{lin2026usergoverned,
  author  = {Jiacheng Lin and Kun Qian and Arvind Srinivasan and Tian Wang and Fang Han and Changran Hu and Junze Liu and Ziyi Wang and Hanwen Xu and Mengmeng Xue and Shuo Yang and Hansi Zeng and Simon Sinong Zhan and Kai Zhong and Weiqi Zhang and Dakuo Wang and Tianhao Wang and Zhiyuan Li},
  title   = {{LLM} Agents Enable User-Governed Personalization Beyond Platform Boundaries},
  howpublished = {arXiv preprint arXiv:2605.09794},
  year    = {2026},
  doi     = {10.48550/arXiv.2605.09794},
  url     = {https://arxiv.org/abs/2605.09794}
}

@inproceedings{zhang2024agentcf,
  author    = {Junjie Zhang and Yupeng Hou and Ruobing Xie and Wenqi Sun and Julian J. McAuley and Wayne Xin Zhao and Leyu Lin and Ji-Rong Wen},
  title     = {{AgentCF}: Collaborative Learning with Autonomous Language Agents for Recommender Systems},
  booktitle = {Proceedings of the ACM Web Conference 2024},
  address   = {Singapore},
  pages     = {3679--3689},
  year      = {2024},
  publisher = {Association for Computing Machinery},
  doi       = {10.1145/3589334.3645537}
}

@misc{zhang2024rec4agentverse,
  author  = {Jizhi Zhang and Keqin Bao and Wenjie Wang and Yang Zhang and Wentao Shi and Wanhong Xu and Fuli Feng and Tat-Seng Chua},
  title   = {Prospect Personalized Recommendation on Large Language Model-based Agent Platform},
  howpublished = {arXiv preprint arXiv:2402.18240},
  year    = {2024},
  doi     = {10.48550/arXiv.2402.18240},
  url     = {https://arxiv.org/abs/2402.18240}
}

@inproceedings{jin2025recinter,
  author    = {Song Jin and Juntian Zhang and Yuhan Liu and Xun Zhang and Yufei Zhang and Guojun Yin and Fei Jiang and Wei Lin and Rui Yan},
  title     = {Beyond Static Testbeds: An Interaction-Centric Agent Simulation Platform for Dynamic Recommender Systems},
  booktitle = {Proceedings of the 2025 Conference on Empirical Methods in Natural Language Processing},
  address   = {Suzhou, China},
  pages     = {18903--18920},
  year      = {2025},
  publisher = {Association for Computational Linguistics},
  doi       = {10.18653/v1/2025.emnlp-main.956},
  url       = {https://aclanthology.org/2025.emnlp-main.956/}
}

@misc{gong2026trirec,
  author  = {Yaxin Gong and Chongming Gao and Chenxiao Fan and Haoyan Liu and Wenjie Wang and Jianshan Sun and Yangyang Li and Fuli Feng and Xiangnan He},
  title   = {Breaking User-Centric Agency: A Tri-Party Framework for Agent-Based Recommendation},
  howpublished = {arXiv preprint arXiv:2603.10673},
  year    = {2026},
  doi     = {10.48550/arXiv.2603.10673},
  url     = {https://arxiv.org/abs/2603.10673}
}

@article{abdollahpouri2020multistakeholder,
  author  = {Himan Abdollahpouri and Gediminas Adomavicius and Robin Burke and Ido Guy and Dietmar Jannach and Toshihiro Kamishima and Jan Krasnodebski and Luiz Pizzato},
  title   = {Multistakeholder Recommendation: Survey and Research Directions},
  journal = {User Modeling and User-Adapted Interaction},
  volume  = {30},
  number  = {1},
  pages   = {127--158},
  year    = {2020},
  doi     = {10.1007/s11257-019-09256-1}
}

@inproceedings{singh2018fairness,
  author    = {Ashudeep Singh and Thorsten Joachims},
  title     = {Fairness of Exposure in Rankings},
  booktitle = {Proceedings of the 24th ACM SIGKDD International Conference on Knowledge Discovery \& Data Mining},
  address   = {London, United Kingdom},
  pages     = {2219--2228},
  year      = {2018},
  publisher = {Association for Computing Machinery},
  doi       = {10.1145/3219819.3220088}
}

@inproceedings{curmei2021quantifying,
  author    = {Mihaela Curmei and Sarah Dean and Benjamin Recht},
  title     = {Quantifying Availability and Discovery in Recommender Systems via Stochastic Reachability},
  booktitle = {Proceedings of the 38th International Conference on Machine Learning},
  address   = {Virtual Event},
  series    = {Proceedings of Machine Learning Research},
  volume    = {139},
  pages     = {2265--2275},
  year      = {2021},
  publisher = {PMLR},
  url       = {https://proceedings.mlr.press/v139/curmei21a.html}
}

@inproceedings{benporat2018game,
  author    = {Omer Ben-Porat and Moshe Tennenholtz},
  title     = {A Game-Theoretic Approach to Recommendation Systems with Strategic Content Providers},
  booktitle = {Advances in Neural Information Processing Systems},
  address   = {Montr{\'e}al, Canada},
  volume    = {31},
  pages     = {1118--1128},
  year      = {2018},
  publisher = {Curran Associates, Inc.},
  url       = {https://proceedings.neurips.cc/paper/2018/hash/a9a1d5317a33ae8cef33961c34144f84-Abstract.html}
}

@inproceedings{yao2023howbad,
  author    = {Fan Yao and Chuanhao Li and Denis Nekipelov and Hongning Wang and Haifeng Xu},
  title     = {How Bad is Top-$K$ Recommendation under Competing Content Creators?},
  booktitle = {Proceedings of the 40th International Conference on Machine Learning},
  address   = {Honolulu, Hawaii, USA},
  series    = {Proceedings of Machine Learning Research},
  volume    = {202},
  pages     = {39674--39701},
  year      = {2023},
  publisher = {PMLR},
  url       = {https://proceedings.mlr.press/v202/yao23b.html}
}

@inproceedings{jagadeesan2023supply,
  author    = {Meena Jagadeesan and Nikhil Garg and Jacob Steinhardt},
  title     = {Supply-Side Equilibria in Recommender Systems},
  booktitle = {Advances in Neural Information Processing Systems},
  address   = {New Orleans, Louisiana, USA},
  volume    = {36},
  pages     = {14597--14608},
  year      = {2023},
  publisher = {Curran Associates, Inc.},
  url       = {https://proceedings.neurips.cc/paper_files/paper/2023/hash/2f1486343c2c942a617e4f5bb0cc64c8-Abstract-Conference.html}
}

@article{jagadeesan2023competition,
  author  = {Meena Jagadeesan and Michael I. Jordan and Nika Haghtalab},
  title   = {Competition, Alignment, and Equilibria in Digital Marketplaces},
  journal = {Proceedings of the AAAI Conference on Artificial Intelligence},
  volume  = {37},
  number  = {5},
  pages   = {5689--5696},
  year    = {2023},
  doi     = {10.1609/aaai.v37i5.25706}
}

@article{tintarev2012evaluating,
  author  = {Nava Tintarev and Judith Masthoff},
  title   = {Evaluating the Effectiveness of Explanations for Recommender Systems},
  journal = {User Modeling and User-Adapted Interaction},
  volume  = {22},
  number  = {4--5},
  pages   = {399--439},
  year    = {2012},
  doi     = {10.1007/s11257-011-9117-5}
}

@inproceedings{qin2024beyond,
  author    = {Peixin Qin and Chen Huang and Yang Deng and Wenqiang Lei and Tat-Seng Chua},
  title     = {Beyond Persuasion: Towards Conversational Recommender System with Credible Explanations},
  booktitle = {Findings of the Association for Computational Linguistics: EMNLP 2024},
  address   = {Miami, Florida, USA},
  pages     = {4264--4282},
  year      = {2024},
  publisher = {Association for Computational Linguistics},
  doi       = {10.18653/v1/2024.findings-emnlp.247},
  url       = {https://aclanthology.org/2024.findings-emnlp.247/}
}

@article{rahman2026persuasive,
  author    = {S. M. Tahsinur Rahman and Dominik Siemon and Tuukka Ruotsalo},
  title     = {Persuasive Explanations for Recommender Systems: How Explanations Can Influence Users' Choices?},
  journal   = {International Journal of Human-Computer Studies},
  volume    = {208},
  articleno = {103720},
  numpages  = {22},
  year      = {2026},
  doi       = {10.1016/j.ijhcs.2025.103720}
}

@misc{allouah2025agentbuying,
  author  = {Amine Allouah and Omar Besbes and Josu{\'e} D. Figueroa and Yash Kanoria and Akshit Kumar},
  title   = {What Is Your {AI} Agent Buying? Evaluation, Biases, Model Dependence, \& Emerging Implications for Agentic E-Commerce},
  howpublished = {arXiv preprint arXiv:2508.02630},
  year    = {2025},
  doi     = {10.48550/arXiv.2508.02630},
  url     = {https://arxiv.org/abs/2508.02630}
}

@article{dellarocas2003digitization,
  author  = {Chrysanthos Dellarocas},
  title   = {The Digitization of Word of Mouth: Promise and Challenges of Online Feedback Mechanisms},
  journal = {Management Science},
  volume  = {49},
  number  = {10},
  pages   = {1407--1424},
  year    = {2003},
  doi     = {10.1287/mnsc.49.10.1407.17308}
}

@article{resnick2006value,
  author  = {Paul Resnick and Richard Zeckhauser and John Swanson and Kate Lockwood},
  title   = {The Value of Reputation on {eBay}: A Controlled Experiment},
  journal = {Experimental Economics},
  volume  = {9},
  number  = {2},
  pages   = {79--101},
  year    = {2006},
  doi     = {10.1007/s10683-006-4309-2}
}

@inproceedings{hou2026bridging,
  author    = {Yupeng Hou and Jiacheng Li and Xiangjun Fu and Zhankui He and An Yan and Xiusi Chen and Julian McAuley},
  title     = {Bridging Language and Items for Retrieval and Recommendation: Benchmarking {LLM}s as Semantic Encoders},
  booktitle = {Proceedings of the 64th Annual Meeting of the Association for Computational Linguistics (Volume 1: Long Papers)},
  address   = {San Diego, California, USA},
  pages     = {3251--3265},
  year      = {2026},
  month     = jul,
  publisher = {Association for Computational Linguistics},
  doi       = {10.18653/v1/2026.acl-long.147},
  url       = {https://aclanthology.org/2026.acl-long.147/}
}

@article{broder2002taxonomy,
  author    = {Andrei Broder},
  title     = {A Taxonomy of Web Search},
  journal   = {ACM SIGIR Forum},
  volume    = {36},
  number    = {2},
  pages     = {3--10},
  year      = {2002},
  publisher = {Association for Computing Machinery},
  doi       = {10.1145/792550.792552}
}

@inproceedings{rose2004understanding,
  author    = {Daniel E. Rose and Danny Levinson},
  title     = {Understanding User Goals in Web Search},
  booktitle = {Proceedings of the 13th International Conference on World Wide Web},
  address   = {New York, NY, USA},
  pages     = {13--19},
  year      = {2004},
  publisher = {Association for Computing Machinery},
  doi       = {10.1145/988672.988675}
}

@article{marchionini2006exploratory,
  author    = {Gary Marchionini},
  title     = {Exploratory Search: From Finding to Understanding},
  journal   = {Communications of the ACM},
  volume    = {49},
  number    = {4},
  pages     = {41--46},
  year      = {2006},
  publisher = {Association for Computing Machinery},
  doi       = {10.1145/1121949.1121979}
}

@inproceedings{white2009expertise,
  author    = {Ryen W. White and Susan T. Dumais and Jaime Teevan},
  title     = {Characterizing the Influence of Domain Expertise on Web Search Behavior},
  booktitle = {Proceedings of the Second ACM International Conference on Web Search and Data Mining},
  address   = {Barcelona, Spain},
  pages     = {132--141},
  year      = {2009},
  publisher = {Association for Computing Machinery},
  doi       = {10.1145/1498759.1498819}
}

@article{bettman1998constructive,
  author  = {James R. Bettman and Mary Frances Luce and John W. Payne},
  title   = {Constructive Consumer Choice Processes},
  journal = {Journal of Consumer Research},
  volume  = {25},
  number  = {3},
  pages   = {187--217},
  year    = {1998},
  doi     = {10.1086/209535}
}

@article{babin1994work,
  author  = {Barry J. Babin and William R. Darden and Mitch Griffin},
  title   = {Work and/or Fun: Measuring Hedonic and Utilitarian Shopping Value},
  journal = {Journal of Consumer Research},
  volume  = {20},
  number  = {4},
  pages   = {644--656},
  year    = {1994},
  doi     = {10.1086/209376}
}

@article{chen2012critiquing,
  author  = {Li Chen and Pearl Pu},
  title   = {Critiquing-Based Recommenders: Survey and Emerging Trends},
  journal = {User Modeling and User-Adapted Interaction},
  volume  = {22},
  number  = {1--2},
  pages   = {125--150},
  year    = {2012},
  doi     = {10.1007/s11257-011-9108-6}
}

@article{friestad1994persuasion,
  author  = {Marian Friestad and Peter Wright},
  title   = {The Persuasion Knowledge Model: How People Cope with Persuasion Attempts},
  journal = {Journal of Consumer Research},
  volume  = {21},
  number  = {1},
  pages   = {1--31},
  year    = {1994},
  doi     = {10.1086/209380}
}

@article{eisend2006twosided,
  author  = {Martin Eisend},
  title   = {Two-Sided Advertising: A Meta-Analysis},
  journal = {International Journal of Research in Marketing},
  volume  = {23},
  number  = {2},
  pages   = {187--198},
  year    = {2006},
  doi     = {10.1016/j.ijresmar.2005.11.001}
}

@misc{ftc1984substantiation,
  author       = {{Federal Trade Commission}},
  title        = {{FTC} Policy Statement Regarding Advertising Substantiation},
  year         = {1984},
  month        = nov,
  url          = {https://www.ftc.gov/legal-library/browse/ftc-policy-statement-regarding-advertising-substantiation},
  note         = {Issued November 23, 1984; appended to Thompson Medical Co., 104 F.T.C. 648, 839}
}
